%% file: main.tex
\PassOptionsToPackage{table}{xcolor}
\documentclass[10pt]{article}
\usepackage{zsight_preprint}

\input{math_commands.tex}

\usepackage{url}
\usepackage{amsmath,amssymb}
\usepackage{array}
\usepackage{colortbl}
\usepackage{xspace}
\usepackage{needspace}
\usepackage{adjustbox}
\usepackage{etoc}
\AtBeginDocument{\etocdepthtag.toc{mtmain}}
\setcitestyle{round,authoryear}

\newcommand{\method}{$\mathcal{Z}$-sight\xspace}
\newcommand{\pizerofive}{$\pi_{0.5}$\xspace}
\newcommand{\ablcell}[2]{#1\textsubscript{\scriptsize(\ensuremath{\downarrow}\,#2\%)}}

\definecolor{zsightteal}{HTML}{0B7D74}
\definecolor{heatblue}{RGB}{76,114,176}
\definecolor{heathead}{gray}{0.95}
\newcommand{\heatmax}{35}
\makeatletter
\def\heat@dash{--}
\newcommand{\heat@cell}[2]{%
  \def\heat@arg{#2}%
  \ifx\heat@arg\heat@dash
    \textcolor{black!35}{--}%
  \else
    \cellcolor{heatblue!\fpeval{round(\heatmax*(#2)/100)}!white}#1{#2}%
  \fi}
\newcommand{\heat}[1]{\heat@cell{}{#1}}
\newcommand{\heatb}[1]{\heat@cell{\bfseries}{#1}}
\makeatother

\headerlogos{%
  \raisebox{1.6pt}{\includegraphics[height=15.2pt]{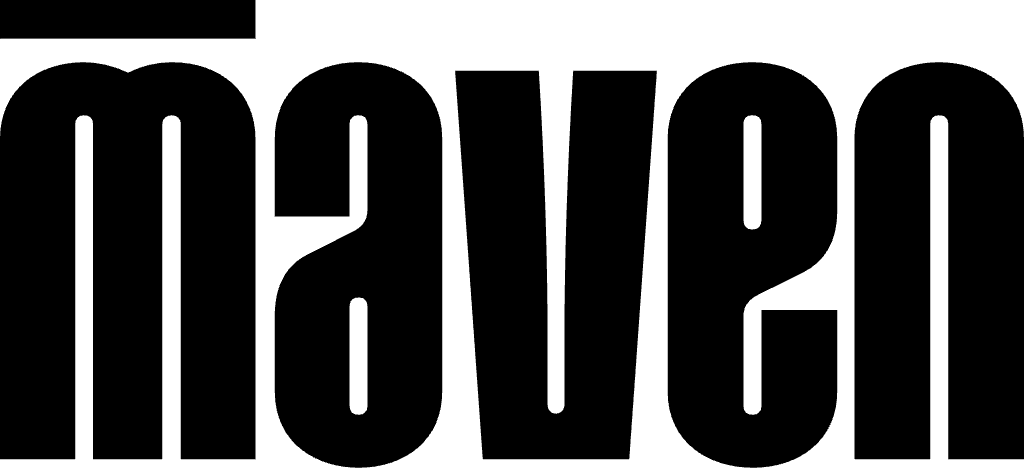}}%
  \hspace{12pt}%
  \raisebox{0.5pt}{\includegraphics[height=15.5pt]{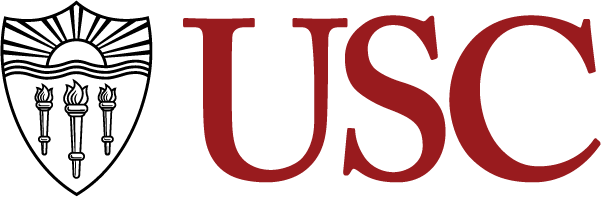}}%
}
\headershorttitle{\raisebox{2pt}{{\normalsize\sffamily\bfseries\textcolor{accentlink}{$\mathcal{Z}$}\textcolor{inkdark}{-sight}}}}

\title{\method: Unifying Vision-Language-Action and\\Latent World Modeling}

\author[1,2,*]{Junjie Ye}
\author[2,*]{Yiqi Zhao}
\author[2,*]{Muchen Xu}
\author[2]{Celina Shiyu Wang}
\author[2]{Hao Dong}
\author[2]{Ruohai Ge}
\author[2]{Rong Xue}
\author[1]{Benjamin J Kolligs}
\author[1]{Aditya Dutt}
\author[1]{Sasanka Nagavalli}
\author[1]{Nassim A. Samad}
\author[2]{Gaurav S. Sukhatme}
\author[2]{Jyotirmoy V. Deshmukh}
\author[2,\dagger]{Yue Wang}
\author[1,\dagger]{Yuliang Guo}

\affiliation[1]{Maven Robotics}
\affiliation[2]{University of Southern California}
\contribution[*]{Equal Contribution}
\contribution[\dagger]{Equal Advising}

\abstract{\input{sec/0_abstract}}

\website{\textcolor{accentlink}{\url{https://junjieye.com/Z-sight/}}} % Replace with \url{https://...}

\begin{document}

\maketitle

\input{sec/1_intro}
\input{sec/2_related_work}
\input{sec/3_method}
\input{sec/4_experiments}
\input{sec/5_conclusion}
\input{sec/XX_acknowledgement}

\clearpage
\bibliographystyle{ieeenat_fullname}
\bibliography{main}

\clearpage
\beginappendix
\input{sec/X_suppl}

\end{document}

%% file: math_commands.tex
\usepackage{amsmath,amsfonts,bm}

\def\eqref#1{equation~\ref{#1}}
\def\1{\bm{1}}

\def\va{{\bm{a}}}

\def\ve{{\bm{e}}}
\def\vf{{\bm{f}}}

\def\vh{{\bm{h}}}

\def\vk{{\bm{k}}}

\def\vq{{\bm{q}}}

\def\vs{{\bm{s}}}

\def\vu{{\bm{u}}}
\def\vv{{\bm{v}}}

\def\vy{{\bm{y}}}
\def\vz{{\bm{z}}}

\def\mA{{\bm{A}}}

\def\mI{{\bm{I}}}

\def\mR{{\bm{R}}}

\def\mW{{\bm{W}}}

\DeclareMathAlphabet{\mathsfit}{\encodingdefault}{\sfdefault}{m}{sl}
\SetMathAlphabet{\mathsfit}{bold}{\encodingdefault}{\sfdefault}{bx}{n}

\def\gL{{\mathcal{L}}}

\def\gN{{\mathcal{N}}}

\newcommand{\E}{\mathbb{E}}

\newcommand{\R}{\mathbb{R}}

%% file: sec/0_abstract.tex
We introduce \method, a unified robot foundation model that couples
vision-language-action (VLA) and generative latent world modeling. Video-based
world action models (WAMs) show the value of joint future-and-action generation
for control, yet inherit the heavy cost of video generators. Recent
latent-space policies avoid pixel generation, yet typically align intermediate
features or rely on frozen target encoders, decoupling future prediction from
the VLM's native feature space. \method
instead predicts the future directly in the visual token space in which its VLM
perceives the present. Coupling a
foresight expert and a continuous action expert with the VLM in a
mixture-of-transformers, \method jointly denoises future spatial tokens and
actions at inference without pixel reconstruction or external encoders. Trained
on human video and robot trajectories, \method achieves 81.9\% success on
SIMPLE (+9.6 points over the strongest baseline) and 73.3\% across six Unitree
G1 tasks (+20.0 points) at 110\,ms per chunk, while remaining competitive on
tabletop benchmarks.

%% file: sec/1_intro.tex
\section{Introduction}
\label{sec:intro}

Vision-language-action (VLA) models transfer visual and linguistic knowledge to
robotic control by building policies on pre-trained VLMs~\citep{rt2, kim2024openvla, pi0}. Recent architectures extend this
paradigm by coupling a VLM backbone with a continuous action expert~\citep{pi05,
gr00tn1_2025, psi02026}, combining multimodal reasoning with high-frequency
action generation.

In parallel, world action models (WAMs)~\citep{dreamzero, motus, fastwam,
rollingwam2026} adapt pre-trained video generators to jointly model future
observations and actions, capturing physical dynamics at the cost of large
video diffusion backbones. Recent methods avoid pixel generation by moving
future supervision into latent space, yet either align intermediate policy
features to compact embeddings~\citep{flare2025, spatialforcing2026} without
generating a future stream, or predict features of frozen target
encoders~\citep{dreamvla2025, vlajepa2026, unijepa2025} that decouple from the
VLM's native feature space.

\begin{figure}[!t]
\centering
\includegraphics[width=0.95\linewidth]{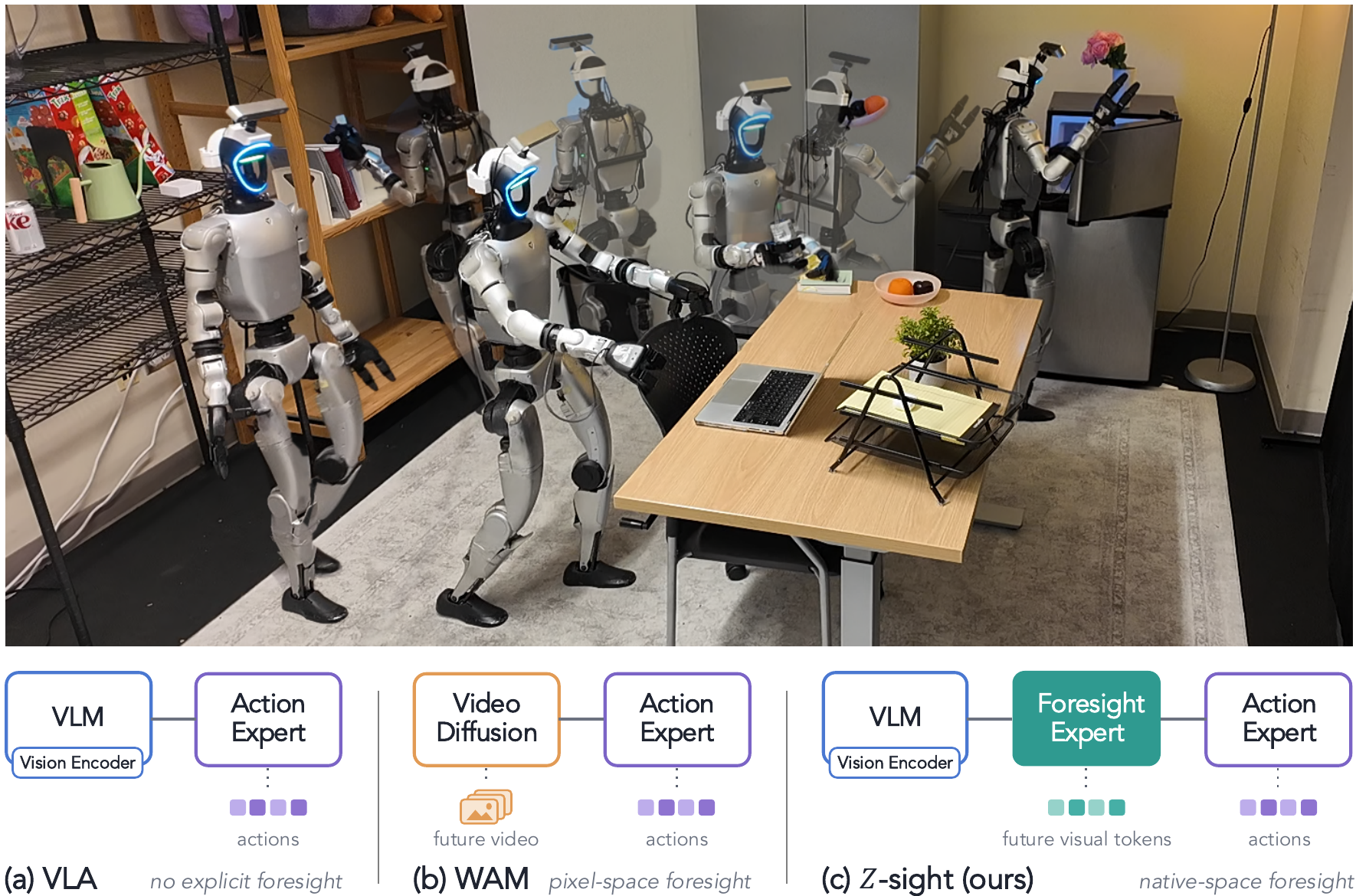}
\caption{\textbf{\method predicts the future in the policy's native representation space.}
\textit{Top:} \method performs diverse loco-manipulation tasks in a real-world setting.
\textit{Bottom:} comparison of policy designs. (a) Vision-language-action
models (VLAs) map the current observation to actions through a VLM and an
action expert. (b) World action models (WAMs) build on a pretrained video
generator to predict future video alongside actions. (c) \method (ours)
co-denoises future spatial tokens and actions in the co-evolving visual space
of its own VLM, allowing the action expert to read the forming prediction
during generation.}
\label{fig:teaser}
\end{figure}

We introduce \method, a unified model that brings joint future-and-action
generation into the policy's co-evolving native visual space
(Figure~\ref{fig:teaser}). A VLM encodes observations into spatial visual
tokens. By supervising future predictions against an exponential-moving-average
(EMA) teacher of this same vision encoder, perception and prediction remain
aligned throughout pre-training and downstream adaptation, without introducing
a video generator or an external target encoder.

As shown in Figure~\ref{fig:method}, \method couples the VLM stream with a
foresight expert and a continuous action expert in a mixture-of-transformers
(MoT)~\citep{mixtureoftransformers}. Conditioned on the VLM's instruction,
state, and observation history, the foresight expert predicts future spatial
tokens via flow matching while the action expert attends directly to these
predicted features. During inference, both experts jointly denoise future
visual tokens and actions, informing control with online latent dynamics at
high efficiency.

We train \method on egocentric human video and cross-embodiment robot
trajectories, then adapt it across single-arm, bimanual, and humanoid settings
spanning LIBERO~\citep{libero}, RoboTwin 2.0~\citep{robotwin},
SIMPLE~\citep{wei2026simple}, and six physical Unitree G1 tasks. \method
achieves strong humanoid loco-manipulation performance, reaching 81.9\% success
on SIMPLE and 73.3\% on the real robot (+9.6 and +20.0 points over the
strongest baselines) at 110\,ms per action chunk, while remaining competitive
on tabletop manipulation benchmarks. Our contributions are
threefold:

\begin{itemize}
    \item We introduce \method, a unified VLA and latent world model that
    jointly denoises future spatial visual tokens and continuous actions within
    the VLM's co-evolving visual representation space.
    \item Through matched architectural controls and test-time interventions,
    we demonstrate the advantages of co-evolving native targets and joint
    denoising over implicit feature alignment and frozen external targets.
    \item \method achieves strong humanoid loco-manipulation results on SIMPLE and six real-world
    Unitree G1 tasks at 110\,ms per chunk, while
    remaining competitive on tabletop manipulation benchmarks.
\end{itemize}

%% file: sec/2_related_work.tex
\section{Related Work}
\label{sec:related_work}

\noindent\textbf{Vision-language-action models.}
Vision-language-action (VLA) models adapt pre-trained VLMs to output discrete
action tokens~\citep{rt2, kim2024openvla, pi0, pertsch2025fast} or continuous
actions via flow-matching experts~\citep{pi05, gr00tn1_2025, psi02026}. Leveraging internet-scale
multimodal pre-training, these policies exhibit strong semantic reasoning and
instruction following. To preserve pre-trained representations while learning
continuous control, \citet{driess2025ki} employ knowledge insulation, training
the VLM with discrete action tokens while a flow expert learns continuous
actions from detached VLM features. While effective at mapping current context
to actions, standard VLAs rely primarily on action labels, providing a
relatively sparse supervision signal. \method builds upon this foundation and
augments it with generative latent world modeling.

\noindent\textbf{World action models and unified models.}
World action models (WAMs) capture physical dynamics by predicting future video
alongside actions, either jointly denoising video and actions on video
diffusion backbones~\citep{dreamzero, rollingwam2026} or using video
co-training while skipping test-time generation~\citep{fastwam}. Motivated by this complementarity,
unified models combine a VLM with visual generation and action experts. For
example, Motus~\citep{motus} couples a VLM, a video generation expert, and an
action expert, while BagelVLA~\citep{hu2026bagelvla} interleaves textual plans,
generated keyframes, and actions with single-step guidance. Related designs
include UP-VLA~\citep{upvla2025}, dual-stream VLAs~\citep{dualstream2025},
F1~\citep{f1vla2025}, InternVLA-A1~\citep{internvlaa1}, and
MotuBrain~\citep{motubrain}. However, relying on pixel or codec generation introduces
substantial computational cost. \method instead performs joint
future-and-action generation directly in the VLM's visual token space without a
video generator.

\noindent\textbf{Latent world modeling and predictive policies.}
Joint-embedding predictive architectures learn representations by predicting
features rather than pixels~\citep{baevski2022data2vec, assran2023ijepa,
bardes2024vjepa} against EMA teachers~\citep{grill2020byol}, enabling planning
with pre-trained visual features~\citep{zhou2024dinowm, assran2025vjepa2}. In
robotic control~\citep{lda1b2026, reconstruction2026}, implicit alignment
methods align internal policy activations with compact future
embeddings~\citep{flare2025, gr00tn15} or spatial and world-model
features~\citep{spatialforcing2026, dao2026thaw} without
generating a spatial future stream. Explicit latent predictors instead
supervise world tokens with frozen external teachers in
DreamVLA~\citep{dreamvla2025}, or frozen V-JEPA~2 and SigLIP encoders in
VLA-JEPA~\citep{vlajepa2026} and UniJEPA~\citep{unijepa2025}. Instead, \method
models future states as dense spatial tokens in the co-evolving perceptual
space of the policy's own vision encoder, jointly denoising future tokens and
continuous actions via flow matching (Appendix~\ref{app:design_axes}).

%% file: sec/3_method.tex
\section{Methodology}
\label{sec:method}

\begin{figure}[!t]
    \centering
    \includegraphics[width=0.95\linewidth]{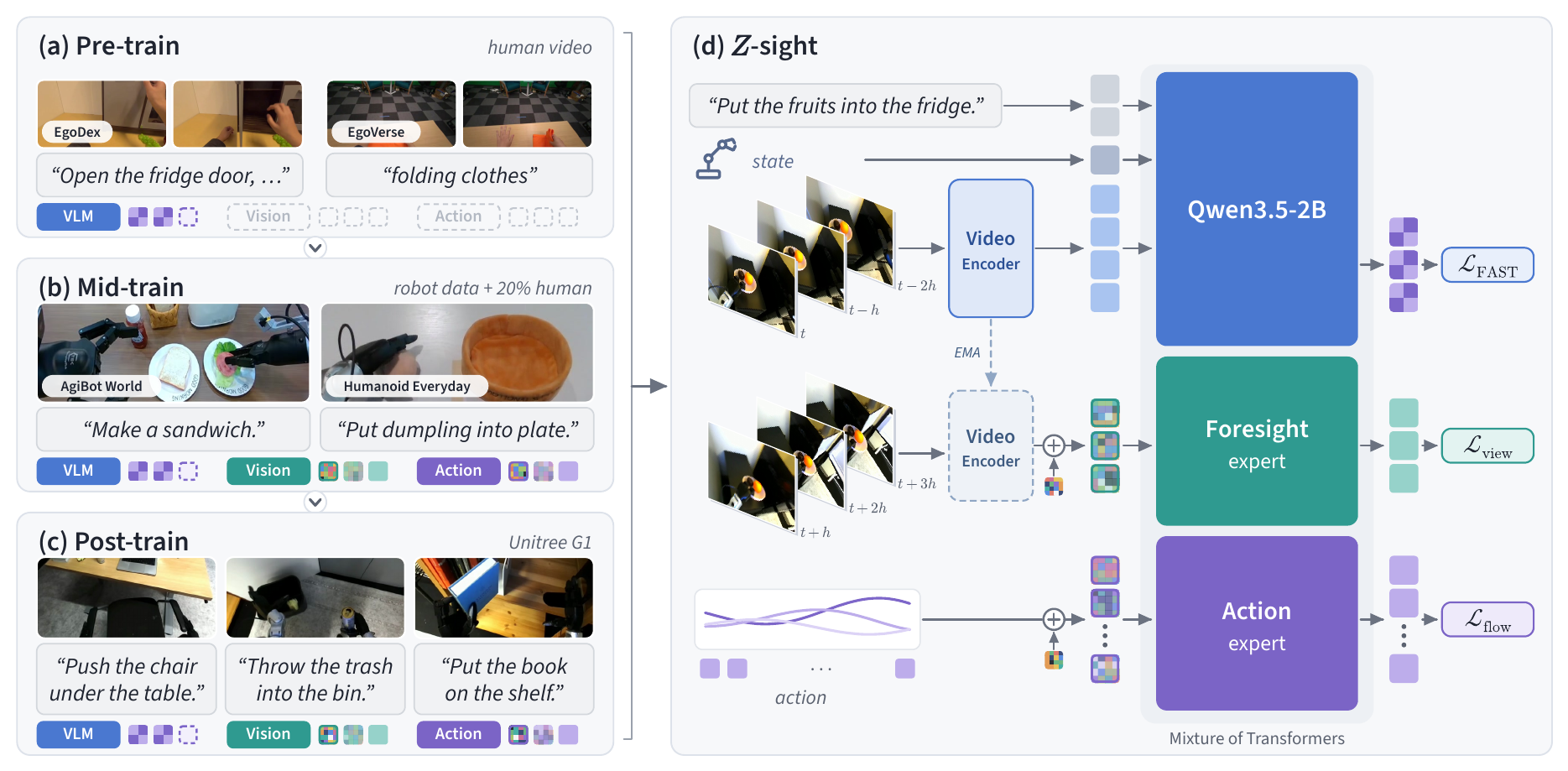}
    \caption{\textbf{Overview of \method.} \textbf{(a)--(c) Training stages:}
    VLM pre-training on human video, mid-training all streams on mixed
    robot-human data, and post-training on the target embodiment.
    \textbf{(d) Architecture:} The VLM processes observation history, state,
    and instructions, supervised by discrete FAST tokens ($\gL_{\text{FAST}}$).
    A foresight expert denoises future spatial tokens against an EMA teacher
    ($\gL_{\text{view}}$). Simultaneously, an action expert generates
    continuous actions via flow matching ($\gL_{\text{flow}}$) while attending
    to the predicted future features.}
    \label{fig:framework}\label{fig:method}
\end{figure}

\method predicts the future in the same visual representation space in which it
perceives the present and uses that prediction to generate actions. At time
$t$, given a language instruction $\ell$, proprioceptive state $\vs_t$, and
multi-camera observation history $\mathbf{O}_t$, the policy outputs an action
chunk $\mA_t = (\va_t, \dots, \va_{t+H-1})$. During training, the future
observations within the chunk, $\mathbf{O}^{+}_t$, serve as an additional
supervision signal that is unavailable at test time.

\subsection{Model Architecture}
\label{sec:arch}

\noindent\textbf{Native representation space.} The VLM's vision encoder
$E_\phi$ maps an image $o$ to $P$ spatial tokens $E_\phi(o)\in\R^{P\times D}$
consumed by the language model. We call this token space the policy's
\textit{native representation space}. Instead of synthesizing pixels, \method
predicts the spatial tokens that $E_\phi$ would produce for future
observations. This preserves the spatial layout of the scene without
introducing a pixel decoder or video generator.

\noindent\textbf{Video encoder for history and future.} A single frame lacks
motion cues, so we extend $E_\phi$ into a parameter-free causal video encoder
$V_\phi$ following MEM~\citep{mem2026}. Every fourth ViT block replaces frame-wise attention with causal
temporal attention followed by spatial attention, and only the final frame's
$P$ tokens are passed to the VLM, retaining the single-image token budget
(Appendix~\ref{app:temporal-encoding}). Multiple cameras are encoded
independently with learned view-role embeddings. To construct prediction
targets, an EMA teacher $V_{\bar\phi}$ tracks the online encoder via
\begin{equation}
  \bar\phi \leftarrow \mu\,\bar\phi + (1-\mu)\,\phi,
\end{equation}
with decay $\mu\in[0,1)$. Applying the two encoders to the observation history
and future clip yields
\begin{equation}
  \vz_t = V_\phi(\mathbf{O}_t), \quad \vz^{+}_t = \operatorname{LN}\!\left(\mathrm{sg}\!\left[V_{\bar\phi}(\mathbf{O}^{+}_t)\right]\right),
\end{equation}
where $\vz_t, \vz^{+}_t \in \R^{P\times D}$ are the perceived visual input and
the future target, $\mathrm{sg}[\cdot]$ denotes
stop-gradient, and $\operatorname{LN}(\cdot)$ applies per-token
layer normalization across feature channels to stabilize target scale.

\noindent\textbf{Preventing collapse and forgetting.} Because $\vz_t$ and
$\vz^{+}_t$ share the same encoder family, backpropagating the future-prediction
loss into $V_\phi$ could cause representation collapse and erode the VLM's
pre-trained knowledge. We prevent both issues through two mechanisms. First,
stop-gradient EMA targets~\citep{assran2023ijepa} keep $\vz^{+}_t$ aligned with
the evolving perceptual encoder without allowing the encoder to trivialize its
own prediction target. Second, we block $\gL_{\text{view}}$ gradients from
entering the VLM across all stages. During mid-training, when the newly added
experts are randomly initialized, we apply full knowledge
insulation~\citep{driess2025ki} so both experts attend to the VLM's keys and
values through stop-gradient, while the VLM is adapted by autoregressively
predicting discrete FAST action tokens~\citep{pertsch2025fast} $c_{1:N}$ of
$\mA_t$
\begin{equation}
  \gL_{\text{FAST}} = -\sum_{n=1}^{N} \log p_{\theta}\big(c_n \mid c_{<n}, \ell, \vs_t, \vz_t\big),
\end{equation}
where $p_\theta$ is the VLM's next-token distribution and $\theta$ includes
$\phi$. Thus $V_\phi$ adapts solely through action supervision, while its EMA
teacher supplies co-evolving future targets. Stage-wise gradient paths are
detailed in Appendix~\ref{app:impl}.

\begin{figure}[!t]
   \centering
   \includegraphics[width=0.41\linewidth]{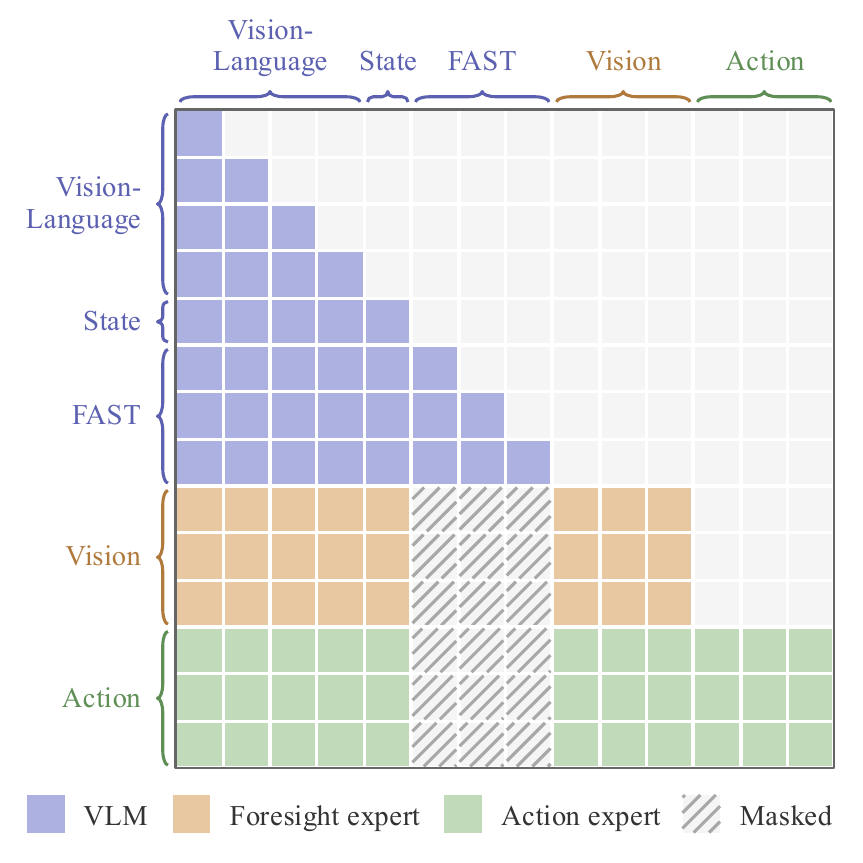}
   \caption{\textbf{Attention mask of \method.}
   The VLM prefix attends causally and cannot see either expert. The foresight
   expert attends to the prefix and to itself. The action expert attends to the
   prefix, the foresight stream, and itself.}
   \label{fig:attn}
\end{figure}

\noindent\textbf{Foresight expert.} Future prediction is performed by a
dedicated foresight expert in the mixture-of-transformers via flow matching in
token space. Given visual noise level $\sigma\in[0,1]$ and
$\boldsymbol{\xi}\sim\gN(\mathbf{0},\mI)$, it receives noisy future tokens
$\vf^{\sigma} = \sigma\boldsymbol{\xi} + (1-\sigma)\,\vz^{+}_t$ combined with
normalized current-frame tokens
\begin{equation}
  \vh = \vf^{\sigma}\mW_{\text{in}} + \mathrm{sg}\!\left[\operatorname{LN}(\vz_t)\right]\mW_{c} + \ve_{\text{pos}},
\end{equation}
where $\mW_{\text{in}}$ and $\mW_c$ project channel dimension $D$ to the expert
width and $\ve_{\text{pos}}$ is a learned position embedding. Zero-initializing $\mW_c$
allows the expert to gradually acquire a direct current-frame conditioning path
alongside VLM attention. Every layer is conditioned on $\sigma$ via adaptive
RMSNorm. Using the $x$-prediction parameterization~\citep{li2025jit}, the
expert predicts clean tokens $\hat{\vz}^{+}_t$ under the implied velocity error
\begin{equation}
  \gL_{\text{view}} = \E\Big[\big\|(\vz^{+}_t - \hat{\vz}^{+}_t)\,/\,\max(\sigma, \sigma_{\min})\big\|^2\Big],
\end{equation}
where $\sigma_{\min}$ bounds the weight at small noise levels.

\noindent\textbf{Action expert.} A continuous action expert generates the
action chunk via conditional flow matching~\citep{flowmatching, pi0}. Given
action noise level $\tau\in[0,1]$ and $\boldsymbol{\epsilon}\sim\gN(\mathbf{0},\mI)$,
it receives $\mA^{\tau} = \tau\boldsymbol{\epsilon} + (1-\tau)\mA_t$, is
conditioned on $\tau$ via adaptive RMSNorm, and predicts velocity $\hat{\vu}_t$
\begin{equation}
  \gL_{\text{flow}} = \E\big[\|\hat{\vu}_t - (\boldsymbol{\epsilon} - \mA_t)\|^2\big].
\end{equation}
During training, we sample $\sigma = \max(\tau, u) \ge \tau$ with a random floor
$u$, exposing the action expert to future streams with varying uncertainty.

\noindent\textbf{Attention pattern and joint denoising.} As shown in
Figure~\ref{fig:attn}, tokens are ordered as
$[\,\texttt{VLM} \mid \texttt{FAST} \mid \texttt{foresight} \mid \texttt{action}\,]$.
VLM and FAST tokens attend causally, and both experts are masked from reading
ground-truth FAST tokens. The foresight stream attends to the VLM prefix and
bidirectionally to itself. The action stream attends to the VLM prefix, the
foresight stream, and itself, reading predicted future tokens directly in the
native visual space. At inference, FAST tokens are omitted, the VLM prefix is
encoded once with cached keys and values, and both experts are integrated
jointly from $\vf^{1},\mA^{1}\sim\gN(\mathbf{0},\mI)$ over $K$ shared Euler
steps ($\Delta=-1/K$, $\sigma=\tau$)
\begin{equation}
  \mA^{\tau+\Delta} = \mA^{\tau} + \Delta\,\hat{\vu}_t, \quad
  \vf^{\tau+\Delta} = \vf^{\tau} + \Delta\,\frac{\vf^{\tau} - \hat{\vz}^{+}_t}{\max(\tau, \sigma_{\min})},
\end{equation}
updating both streams in a single forward pass per step without pixel decoding.

\subsection{Training Recipe}
\label{sec:recipe}

\begin{figure}[!t]
  \centering
  \includegraphics[width=0.42\linewidth]{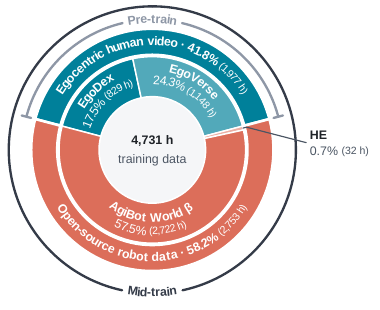}
  \caption{\textbf{Training data.} Pre-training uses 1{,}977\,h of egocentric
  human video (inner arc) and mid-training uses all 4{,}731\,h (outer circle),
  combining robot data with 20\% human-data replay.}
  \label{fig:data}
\end{figure}

We train \method in three stages, comprising VLM pre-training on egocentric
human video, full-model mid-training on cross-embodiment robot data with human
replay, and post-training on the target embodiment (Figure~\ref{fig:data},
Appendix~\ref{app:data}). During pre-training and mid-training, human hands and
humanoids use a 48-D task-space hand representation~\citep{psi02026}, while
other robots use end-effector poses with gripper or hand commands. Actions are
expressed relative to the current state within each chunk and padded to a
shared dimension with per-channel validity masks.

\noindent\textbf{Pre-training on human video.} We train only the VLM with
$\gL_{\text{FAST}}$ on $\sim$2{,}000 hours of video from
EgoDex~\citep{hoque2025egodex} and EgoVerse~\citep{egoverse2026}. Hand trajectories 
are expressed in the current frame to factor out head
motion, low-confidence joints are masked, and trajectories are smoothed with a
validity-weighted temporal Gaussian filter to
suppress frame-to-frame tracking jitter in FAST tokens
(Appendix~\ref{app:smooth}).

\noindent\textbf{Mid-training on cross-embodiment data.} We introduce the
foresight and action experts and optimize
$\gL_{\text{FAST}} + \gL_{\text{flow}} + \lambda_v\gL_{\text{view}}$ under
knowledge insulation on $\sim$2{,}750 hours from AgiBot
World~\citep{agibot2025} and Humanoid Everyday~\citep{zhao2025humanoideveryday}, deriving task-space
poses for joint-only humanoid logs via forward kinematics~\citep{psi02026} and
filtering idle segments. Replaying human video at 20\% of the sampling mass
retains pre-trained priors and supplies diverse visual dynamics.

\noindent\textbf{Post-training on the target embodiment.} On target
demonstrations, we drop $\gL_{\text{FAST}}$ and optimize
$\gL_{\text{flow}} + \lambda_v\gL_{\text{view}}$. Action gradients update the
VLM, state projection, and experts, while $\gL_{\text{view}}$ updates only the
foresight expert with its gradient into
the VLM blocked. The EMA teacher continues tracking the adapting vision
encoder.

%% file: sec/4_experiments.tex
\section{Experiments}
\label{sec:exp}

\method jointly denoises future visual tokens and continuous actions within the
VLM's native feature space. Our experiments evaluate the practical
benefits and underlying mechanisms of this design.
Specifically, we aim to answer four key questions.
(1) How does \method perform compared to strong VLAs and video-based WAMs?
(Section~\ref{sec:main-results})
(2) How much do continued future supervision and the co-evolving native target
design contribute relative to implicit alignment and frozen external encoders?
(Section~\ref{sec:exp-ablation})
(3) What scene changes does the foresight stream capture, and how does the
action expert rely on its content and uncertainty during execution?
(Section~\ref{sec:exp-use})
(4) Can \method retain online foresight at high efficiency?
(Section~\ref{sec:exp-cost})

\subsection{Evaluation Setup}
\noindent\textbf{Benchmarks.} We evaluate \method on three simulation
benchmarks and a physical humanoid. \textbf{SIMPLE}~\citep{wei2026simple}
evaluates whole-body humanoid loco-manipulation across three levels of
variation in materials, distractors, lighting, and target positions
(L0--L2). On a physical \textbf{Unitree G1}, we evaluate six tasks spanning
tabletop manipulation, object transport, and multi-stage loco-manipulation,
reporting success rates and stage-based progress scores
(Appendices~\ref{app:deploy} and~\ref{app:tasks}). \textbf{LIBERO}~\citep{libero}
tests single-arm manipulation across spatial, object,
goal, and long-horizon suites, while \textbf{RoboTwin 2.0}~\citep{robotwin}
evaluates bimanual manipulation over 50 tasks trained on clean demonstrations
and tested under clean (C2C) and randomized (C2R) conditions.

\noindent\textbf{Baselines.} Our primary baselines span representative VLAs
(\pizerofive~\citep{pi05}, GR00T N1.7~\citep{gr00tn17}, and
$\Psi_0$~\citep{psi02026}) and a video-based WAM (Fast-WAM~\citep{fastwam}).
Additional tabletop baselines and adaptation protocols are detailed in
Appendices~\ref{app:baseline} and~\ref{app:simulation}.

\subsection{Performance across Embodiments}\label{sec:main-results}
Because our pre-training and mid-training incorporate human egocentric video
and humanoid trajectories, we first evaluate whole-body humanoid
loco-manipulation in simulation and the real world, then test whether the same
model extends to single-arm and bimanual tabletop manipulation.

\begin{figure}[!t]
  \centering
  \includegraphics[width=0.94\linewidth]{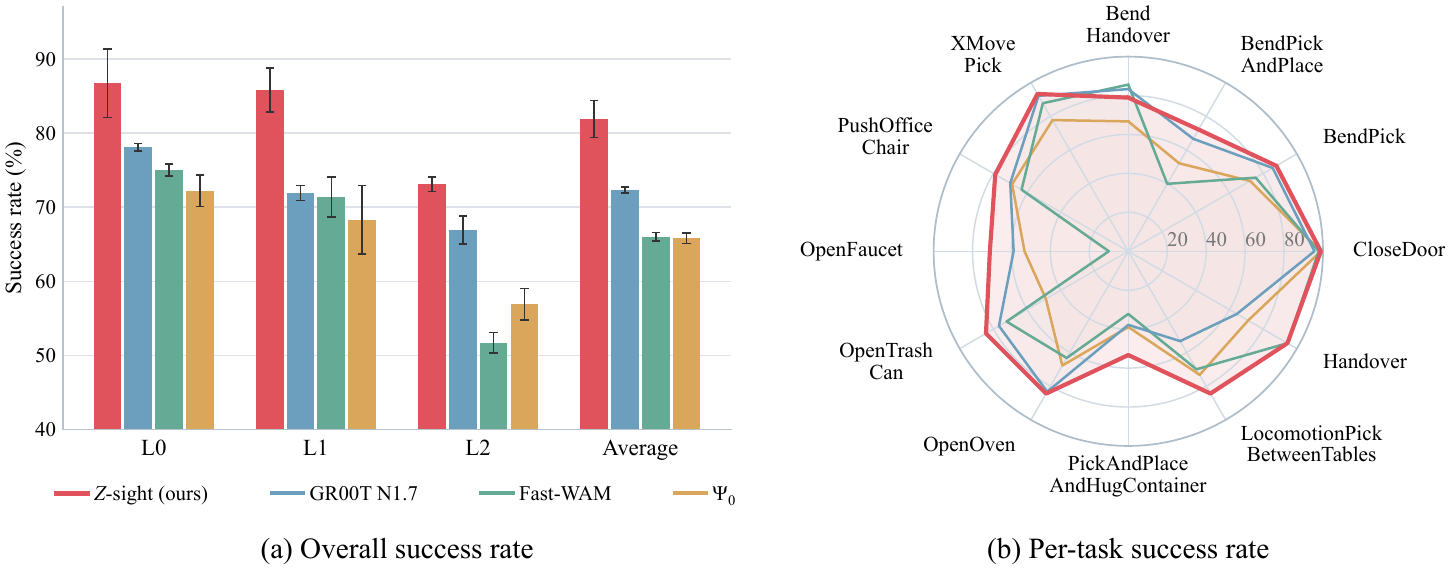}
  \caption{\textbf{Results on SIMPLE.} (a) Success rates by difficulty
  level and overall average. (b) Success rates across the 12 teleoperation tasks
  (averaged over L0--L2). Values are means over three evaluation passes of one
  checkpoint per method, with error bars showing $\pm$ one sample standard
  deviation across passes (ten episodes per task and level per pass).
  \method leads at every difficulty level (81.9\%
  average, +9.6 points over GR00T N1.7) and ranks first or tied on 11 of the 12
  tasks.}
  \label{fig:simple}
\end{figure}

\noindent\textbf{Humanoid loco-manipulation in SIMPLE.}
\label{sec:exp-simple}
All four methods are adapted on the same 17-task SIMPLE dataset and evaluated
over three passes on the 12 teleoperation tasks (full protocol and motion-planned 
task results in Appendix~\ref{app:simple}).
Figure~\ref{fig:simple}a shows that \method achieves 81.9\% overall success,
outperforming GR00T N1.7 by 9.6 points, Fast-WAM by 15.9 points, and $\Psi_0$ 
by 16.1 points, while leading at every
difficulty level (86.7\% on L0, 85.8\% on L1, and 73.1\% on L2). Across
individual tasks in Figure~\ref{fig:simple}b, \method ranks first or tied on 11
of the 12 tasks, with the largest margins on tasks requiring coupled
navigation, whole-body bending, or articulated contact such as
\texttt{LocomotionPickBetweenTables} (84.4\% vs.\ 73.3\% for $\Psi_0$ and
53.3\% for GR00T N1.7), \texttt{PickAndPlaceAndHugContainer} (53.3\% vs.\
38.9\%), and \texttt{OpenFaucet} (71.1\% vs.\ 58.9\%). While action-only VLAs
and test-time foresight-free Fast-WAM degrade sharply under target-position
randomization at L2, co-denoising future spatial tokens helps \method sustain
coordinated whole-body reaching and contact transitions.

\begin{figure}[!t]
  \centering
  \includegraphics[width=\linewidth]{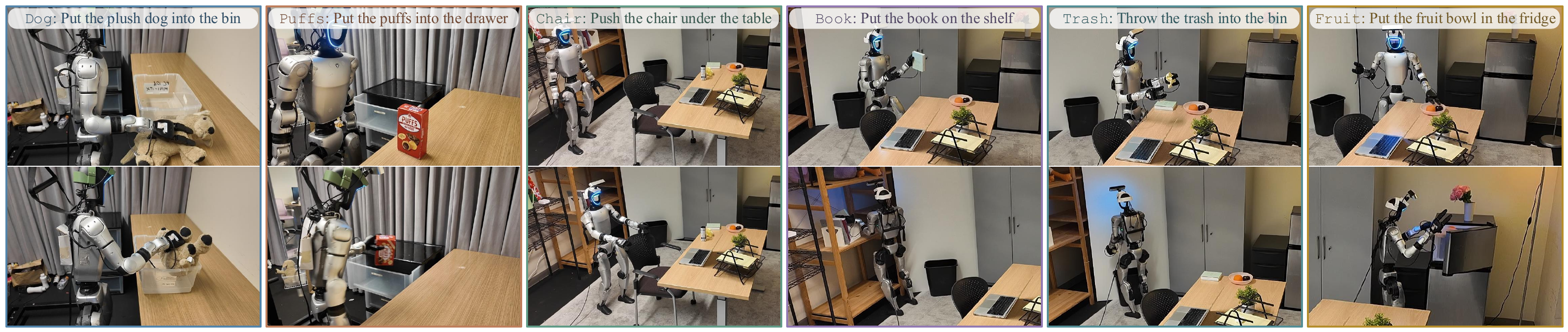}
  \caption{\textbf{Real-world tasks.} The six evaluation tasks on the Unitree
  G1 with task instructions overlaid. \texttt{Dog} and \texttt{Puffs} are
  short pick-and-place tasks, whereas \texttt{Chair}, \texttt{Book},
  \texttt{Trash}, and \texttt{Fruit} are multi-stage loco-manipulation tasks
  requiring coordinated whole-body motion and dexterous grasping.}
  \label{fig:task_setup}
\end{figure}

\begin{figure}[!t]
  \centering
  \includegraphics[width=0.94\linewidth]{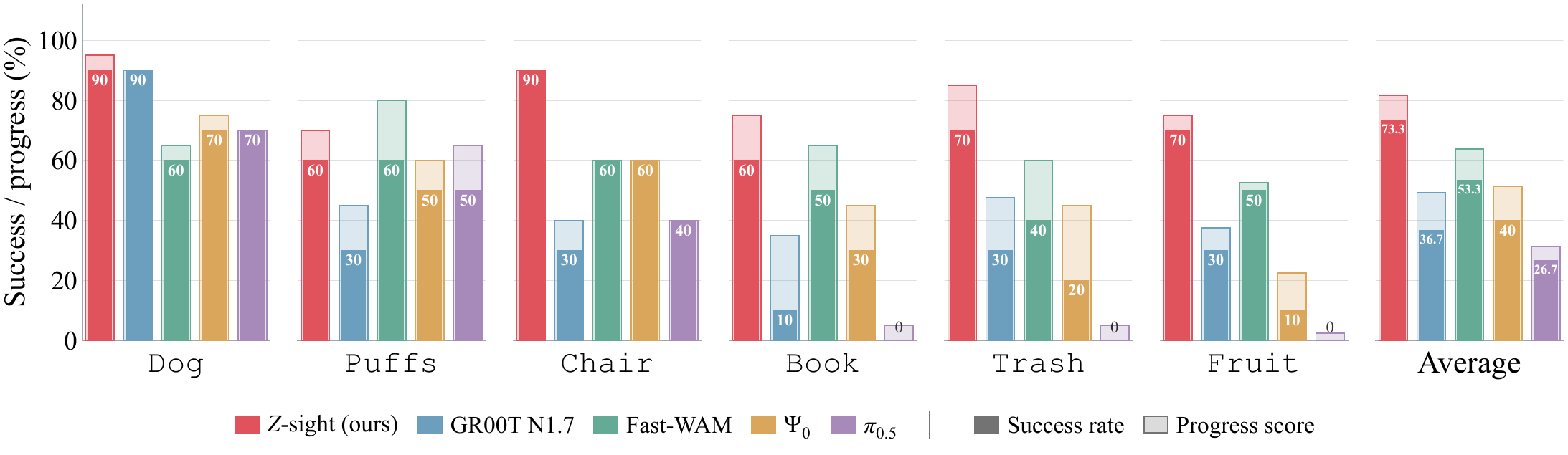}
  \caption{\textbf{Real-world results on the Unitree G1.} Success rate (SR) and
  progress score (\%) over 10 trials per task. Solid bars and labels show SR,
  overlaid on translucent bars showing progress. Average weights all six tasks
  equally. We compare Fast-WAM~\citep{fastwam}, \pizerofive~\citep{pi05}, GR00T
  N1.7~\citep{gr00tn17}, and $\Psi_0$~\citep{psi02026}. \method matches or exceeds
  every baseline on every task, raising overall success to 73.3\% (+20.0 points
  over Fast-WAM) with the largest gains on multi-stage tasks.}
  \label{tab:main}
\end{figure}

\noindent\textbf{Real-world humanoid loco-manipulation.}\label{sec:exp-main}
Figures~\ref{fig:task_setup} and~\ref{tab:main} present our evaluation across six
everyday tasks on a physical Unitree G1 with 10 trials per task. \method reaches 73.3\%
average success (44/60 trials) and 81.7\% progress, outperforming the strongest
baseline Fast-WAM (53.3\% success, 63.8\% progress) as well as $\Psi_0$
(40.0\%), GR00T N1.7 (36.7\%), and \pizerofive (26.7\%), while matching or
exceeding every baseline on every task. The advantage is most pronounced on the
four multi-stage tasks (\texttt{Chair}, \texttt{Book}, \texttt{Trash}, and
\texttt{Fruit}), where \method averages 72.5\% success compared with 50.0\% for
Fast-WAM and at most 32.5\% for the VLA baselines, completing both four-stage
tasks (\texttt{Trash} and \texttt{Fruit}) in 7 of 10 trials. Whereas VLA
baselines frequently stall after completing the initial subgoal, online
native-space foresight helps \method carry sequential stage transitions through
to completion.

\begin{table}[!t]
  \centering
  \begin{minipage}[t]{0.53\linewidth}
    \centering
    \caption{\textbf{LIBERO success rates (\%).} \method outperforms the VLA
    baselines and VLA-JEPA overall, while matching Motus.}
    \label{tab:libero_full}
    \footnotesize
    \setlength{\tabcolsep}{3.8pt}
    \adjustbox{max width=\linewidth}{%
    \begin{tabular}{@{}lccccc@{}}
    \toprule
    Method & Spatial & Object & Goal & Long & Avg. \\
    \midrule
    $\pi_0$~\citep{pi0} & 96.8 & 98.8 & 95.8 & 85.2 & 94.1 \\
    \pizerofive~\citep{pi05} & \textbf{98.8} & 98.2 & \textbf{98.0} & 92.4 & 96.9 \\
    GR00T N1.7~\citep{gr00tn17} & 97.7 & 98.5 & 97.5 & 94.4 & 97.0 \\
    VLA-JEPA~\citep{vlajepa2026} & 96.2 & 99.6 & 97.2 & 95.8 & 97.2 \\
    Motus~\citep{motus} & 96.8 & 99.8 & 96.6 & \textbf{97.6} & \textbf{97.7} \\
    Fast-WAM~\citep{fastwam} & 98.2 & \textbf{100.0} & 97.0 & 95.2 & 97.6 \\
    starVLA~\citep{starvla2026} & 97.8 & 98.6 & 96.2 & 93.8 & 96.6 \\
    \midrule
    \method (ours) & 97.2 & \textbf{100.0} & 97.0 & 96.6 & \textbf{97.7} \\
    \bottomrule
    \end{tabular}}
  \end{minipage}\hfill
  \begin{minipage}[t]{0.44\linewidth}
    \centering
    \caption{\textbf{RoboTwin 2.0.} Success rates (\%) under clean (C2C) and
    randomized (C2R) evaluation.}
    \label{tab:robotwin}
    \footnotesize
    \setlength{\tabcolsep}{4pt}
    \adjustbox{max width=\linewidth}{%
    \begin{tabular}{@{}lrrr@{}}
    \toprule
    Method & C2C & C2R & Avg. \\
    \midrule
    GR00T N1.7~\citep{gr00tn17} & 43.6 & 20.7 & 32.15 \\
    starVLA~\citep{starvla2026} & 46.52 & 3.16 & 24.84 \\
    $\pi_0$~\citep{pi0} & 46.42 & 16.34 & 31.38 \\
    \pizerofive~\citep{pi05} & 70.66 & \textbf{46.00} & 58.33 \\
    Abot-M0~\citep{abotm02026} & 57.40 & 30.36 & 43.88 \\
    Xiaomi Robotics-0~\citep{xiaomirobotics02026} & 62.86 & 18.22 & 40.54 \\
    Fast-WAM~\citep{fastwam} & 77.76 & 1.92 & 39.84 \\
    X-WAM~\citep{xwam2026} & 69.98 & 25.84 & 47.91 \\
    Spatial Forcing~\citep{spatialforcing2026} & 77.18 & 26.74 & 51.96 \\
    \midrule
    \method (ours) & \textbf{82.44} & 39.00 & \textbf{60.72} \\
    \bottomrule
    \end{tabular}}
  \end{minipage}
\end{table}

\noindent\textbf{Single-arm and bimanual tabletop manipulation.}
\label{sec:exp-sim}
Table~\ref{tab:libero_full} reports results on single-arm LIBERO~\citep{libero},
where \method achieves 97.7\% average success. It outperforms the VLA baselines (\pizerofive at
96.9\%, GR00T N1.7 at 97.0\%, and starVLA at 96.6\%) as well as
VLA-JEPA~\citep{vlajepa2026} (97.2\%), with the clearest margin over VLAs on
the long-horizon \texttt{Long} suite (96.6\% vs.\ 92.4--94.4\%), while matching
the video-based Motus average without a video generator.

On bimanual RoboTwin~2.0~\citep{robotwin}, a single RGB-only policy trained on
50 clean demonstrations per task is evaluated over 100 episodes per task under
clean (C2C) and randomized (C2R) conditions. In Table~\ref{tab:robotwin}, \method
reaches 82.44\% on C2C, surpassing all compared VLAs and WAMs including
Fast-WAM (+4.68 points) and Spatial Forcing (+5.26 points). Under randomized
evaluation (C2R), \method reaches 39.00\%, trailing \pizerofive (46.00\%) but
substantially outperforming every other VLA and WAM baseline to achieve
the highest C2C/C2R average (60.72\%). Together,
these benchmarks confirm that \method transfers competitively to standard
tabletop manipulation.

\subsection{Ablating Foresight and Prediction Design}\label{sec:exp-ablation}

\begin{table}[!t]
  \centering
  \caption{\textbf{Training ablations of the foresight stream.} Scores (\%), with
  subscripts showing relative drops from the full model, and gray marking the
  best. Removing either the future-prediction loss or the entire foresight branch
  consistently degrades success across real-world and simulated benchmarks,
  confirming that the gains come from continued predictive supervision rather
  than added model capacity.}
  \label{tab:ablation}
  \small
  \setlength{\tabcolsep}{5.5pt}
  \adjustbox{max width=\linewidth}{%
  \begin{tabular}{l|c|cc|c|cc}
  \toprule[1pt]
  & SIMPLE & \multicolumn{2}{c|}{Real-world \texttt{Dog}} & LIBERO & \multicolumn{2}{c}{RoboTwin 2.0} \\
  Variant & Hard SR & SR & Prog. & Avg. SR & C2C & C2R \\
  \midrule
  \method (ours) & \cellcolor{black!15}73.1 & \cellcolor{black!15}90 & \cellcolor{black!15}95.0 & \cellcolor{black!15}97.70 & \cellcolor{black!15}82.40 & \cellcolor{black!15}40.10 \\
  \midrule
  No view loss & \ablcell{25.0}{65.8} & \ablcell{30}{66.7} & \ablcell{30.0}{68.4} & \ablcell{74.65}{23.6} & \ablcell{60.90}{26.1} & \ablcell{11.60}{71.1} \\
  No foresight branch & \ablcell{68.3}{6.6} & \ablcell{40}{55.6} & \ablcell{40.0}{57.9} & \ablcell{95.35}{2.4} & \ablcell{76.30}{7.4} & \ablcell{22.00}{45.1} \\
  \bottomrule[1pt]
  \end{tabular}}
\end{table}

Starting from the shared mid-trained checkpoint, we first isolate the effect of
continued future-feature supervision during target-domain adaptation, holding
demonstrations and optimization schedules fixed within each benchmark. \textbf{No view loss} keeps the foresight
expert and token stream while setting $\lambda_v=0$, so the branch is updated
only by action-loss gradients. \textbf{No foresight branch} removes the
foresight stream altogether, reducing the policy to a standard VLA.

Across all four settings in Table~\ref{tab:ablation}, removing the view loss consistently degrades performance, dropping success from 73.1\% to 25.0\% on SIMPLE Hard, from 90\% to
30\% on G1 \texttt{Dog}, from 97.70\% to 74.65\% on LIBERO, and from 40.1\% to
11.6\% on RoboTwin C2R. Removing the foresight branch entirely also trails the
full model (e.g., 68.3\% vs.\ 73.1\% on SIMPLE Hard and 22.0\% vs.\ 40.1\% on
RoboTwin C2R). Because the no-view-loss variant preserves parameter count
and architecture, these gains stem from continued predictive supervision rather
than extra model capacity.

\begin{table}[!t]
\centering
\caption{\textbf{Future-prediction designs.} Success rates (\%) on SIMPLE Hard and RoboTwin 2.0. \method
consistently outperforms both implicit feature alignment and frozen external
targets.}
\label{tab:prediction-designs}
\small
\setlength{\tabcolsep}{10pt}
\begin{tabular}{@{}lccc@{}}
\toprule
& SIMPLE & \multicolumn{2}{c}{RoboTwin 2.0} \\
\cmidrule(lr){3-4}
Variant & Hard & C2C & C2R \\
\midrule
FLARE-style~\citep{flare2025} & 67.5 & 74.6 & 20.9 \\
DINOv2-B target~\citep{oquab2023dinov2} & 70.8 & 76.7 & 26.6 \\
\method (ours) & \textbf{73.1} & \textbf{82.4} & \textbf{40.1} \\
\bottomrule
\end{tabular}
\end{table}

\noindent\textbf{Prediction design.} To test whether different forms of latent
future supervision are interchangeable, Table~\ref{tab:prediction-designs} compares two controlled alternatives
from the same mid-trained checkpoint.
\textbf{DINOv2-B target} retains the foresight expert and joint denoising but
replaces our co-evolving native targets with frozen DINOv2-B/14 features.
\textbf{FLARE-style} removes the denoising expert and aligns future-query
tokens on the policy to the same DINOv2-B features via cosine loss. Keeping an
explicit generative foresight stream with frozen DINOv2-B targets improves over
implicit alignment (26.6\% vs.\ 20.9\% on RoboTwin C2R and 70.8\% vs.\ 67.5\%
on SIMPLE Hard). Adopting \method's co-evolving native EMA targets yields a
further substantial jump to 40.1\% on RoboTwin C2R (+13.5 points), 82.4\% on
C2C (+5.7 points), and 73.1\% on SIMPLE Hard, supporting the combination of a
co-evolving native target space and joint generative denoising.

\subsection{Foresight Content and Policy Reliance}\label{sec:exp-use}

\begin{table}[!t]
  \centering
  \begin{minipage}[t]{0.47\linewidth}
    \centering
    \caption{\textbf{Foresight prediction and action quality.} Pooled over
    11{,}676 held-out G1 windows (predicted $\alpha$, present $\beta$, unrelated
    $\gamma$, true future $y$).}
    \label{tab:prediction-quality-main}
    \footnotesize
    \setlength{\tabcolsep}{4pt}
    \adjustbox{max width=\linewidth}{%
    \begin{tabular}{@{}llr@{}}
    \toprule
    Comparison & Metric & Value \\
    \midrule
    Predicted vs.\ future & $\cos(\alpha,y)$ & 0.955 \\
    Present vs.\ future & $\cos(\beta,y)$ & 0.921 \\
    Unrelated vs.\ future & $\cos(\gamma,y)$ & 0.850 \\
    Gain over present & $\cos(\alpha,y)-\cos(\beta,y)$ & $+0.034$ \\
    Change cosine & $\cos(\alpha-\beta,y-\beta)$ & 0.646 \\
    \midrule
    Future-feature input & Action MSE $\downarrow$ & Change \\
    \midrule
    Model prediction $\alpha$ & 0.0495 &  \\
    True future $y$ & 0.0436 & $\downarrow11.9\%$ \\
    \bottomrule
    \end{tabular}}
  \end{minipage}\hfill
  \begin{minipage}[t]{0.50\linewidth}
    \centering
    \caption{\textbf{Future-stream interventions.} Success rates (\%) with fixed
    weights ($\sigma=\tau$: shared schedule; $\sigma=1$: pure noise).}
    \label{tab:content-interventions}
    \footnotesize
    \setlength{\tabcolsep}{4pt}
    \adjustbox{max width=\linewidth}{%
    \begin{tabular}{@{}lcccc@{}}
    \toprule
    & & SIMPLE & \multicolumn{2}{c}{RoboTwin} \\
    \cmidrule(lr){4-5}
    Future stream & Label & Hard & C2C & C2R \\
    \midrule
    Model prediction & $\sigma=\tau$ & 73.1 & 82.4 & 40.1 \\
    Gaussian noise & $\sigma=1$ & 61.7 & 79.0 & 33.9 \\
    Gaussian noise & $\sigma=\tau$ & 16.7 & 50.5 & 15.3 \\
    Swapped prediction & $\sigma=\tau$ & 11.7 & 7.5 & 4.0 \\
    Present frame (Copy) & $\sigma=\tau$ & 5.8 & 7.8 & 2.0 \\
    Attention blocked (Detach) & $\sigma=\tau$ & 7.5 & 5.8 & 1.8 \\
    \bottomrule
    \end{tabular}}
  \end{minipage}
\end{table}

\noindent\textbf{Prediction quality and action error.} On 11{,}676
held-out observation-to-future windows across all six physical G1 tasks,
Table~\ref{tab:prediction-quality-main} compares the predicted future tokens
$\alpha$ against the EMA target encoder's ground-truth future tokens $y$.
\method achieves 0.955 cosine similarity, outperforming both copying present
tokens $\beta$ (0.921) and an unrelated future
$\gamma$ (0.850), while reaching 0.646 cosine similarity on feature changes
$(\alpha-\beta, y-\beta)$. Moreover, holding initial action noise fixed on the
same trajectories, replacing $\alpha$ with ground-truth future tokens $y$
lowers normalized action MSE against demonstrations by
11.9\% (from 0.0495 to 0.0436), confirming that more accurate future features
directly improve sampled actions.

\noindent\textbf{Closed-loop interventions.} To disentangle reliance on
predicted content from the stream's noise-level cue, Table~\ref{tab:content-interventions} evaluates test-time interventions on the
foresight stream with fixed weights, with extended benchmarks in
Appendix~\ref{sec:foresight-analysis}. Injecting Gaussian noise while labeling
the stream unreliable ($\sigma=1$) retains 61.7\% success on SIMPLE Hard and
79.0/33.9\% on RoboTwin, as training with $\sigma\ge\tau$ teaches the policy to
fall back on current observations under high declared uncertainty. Changing
only the label of that noise to the shared schedule $\sigma=\tau$ drops success
to 16.7\% and 50.5/15.3\%. Under the same $\sigma=\tau$ schedule, replacing the
stream with another episode's valid prediction (\textbf{Swap}) preserves token
statistics and the denoising schedule yet collapses success to 11.7\% on SIMPLE
and 7.5/4.0\% on RoboTwin, matching the severe drops from asserting a static
scene (\textbf{Copy}, 5.8\% and 7.8/2.0\%) or blocking action-to-foresight
attention (\textbf{Detach}, 7.5\% and 5.8/1.8\%). Thus, closed-loop control
depends on both declared uncertainty and observation-specific predicted
content.

\begin{table}[!t]
  \centering
  \begin{minipage}[t]{0.54\linewidth}
    \centering
    \caption{\textbf{Joint versus sequential denoising.} Success rates (\%) with
    shared weights and median chunk latency on one RTX 4090 (ms).}
    \label{tab:joint-vs-seq}
    \footnotesize
    \setlength{\tabcolsep}{3.5pt}
    \adjustbox{max width=\linewidth}{%
    \begin{tabular}{@{}lcccccc@{}}
    \toprule
    & & SIMPLE & \multicolumn{2}{c}{RoboTwin} & LIBERO & Latency \\
    \cmidrule(lr){4-5}
    Schedule & Steps & Hard & C2C & C2R & Avg. & (ms) $\downarrow$ \\
    \midrule
    Joint (ours) & 10 & \textbf{73.1} & 82.4 & \textbf{40.1} & \textbf{97.70} & \textbf{114.8} \\
    Sequential   & 20 & 64.2 & \textbf{82.6} & 37.9 & 97.05 & 208.0 \\
    \bottomrule
    \end{tabular}}
  \end{minipage}\hfill
  \begin{minipage}[t]{0.43\linewidth}
    \centering
    \caption{\textbf{Inference latency on one RTX 4090 (ms).}
    \method runs faster than \pizerofive, $\Psi_0$, and Fast-WAM with online foresight.}
    \label{tab:latency}
    \footnotesize
    \setlength{\tabcolsep}{3.8pt}
    \adjustbox{max width=\linewidth}{%
    \begin{tabular}{@{}lccc@{}}
    \toprule
    Method & Median & Mean $\pm$ std & Chunk \\
    \midrule
    \pizerofive~\citep{pi05} & 120.5 & 121.0 $\pm$ 1.3 & 30 \\
    GR00T N1.7~\citep{gr00tn17} & 60.0 & 60.3 $\pm$ 1.4 & 30 \\
    Fast-WAM~\citep{fastwam} & 155.9 & 156.6 $\pm$ 2.1 & 32 \\
    $\Psi_0$~\citep{psi02026} & 149.7 & 149.4 $\pm$ 2.6 & 30 \\
    \midrule
    \method (ours) & 109.5 & 109.6 $\pm$ 0.2 & 30 \\
    \bottomrule
    \end{tabular}}
  \end{minipage}
\end{table}

\noindent\textbf{Joint versus sequential denoising.} Because \method updates
future tokens and actions simultaneously, a natural question is whether
conditioning actions on intermediate noisy predictions hurts control compared
with waiting for a fully denoised future estimate. Table~\ref{tab:joint-vs-seq}
evaluates the same weights under sequential inference, completing 10
future-denoising steps before running 10 action-denoising steps. Sequential
inference yields similar results on LIBERO (97.05\% vs.\ 97.70\%) and RoboTwin
(82.6/37.9\% vs.\ 82.4/40.1\%), but drops SIMPLE Hard success by 8.9 points
from 73.1\% to 64.2\% while increasing median latency by $1.81\times$ from
114.8 to 208.0\,ms on an RTX 4090 (profiling details in
Appendix~\ref{app:latency}). Single-schedule joint denoising is therefore both
faster and at least as effective.

\subsection{Inference Efficiency}
\label{sec:exp-cost}

We time in-process action-chunk generation on a single deployment RTX 4090 over
100 CUDA-synchronized calls on $480\times640$ observations after 100 warm-up
calls (Appendix~\ref{app:latency}). As shown in Table~\ref{tab:latency},
\method computes a 30-step chunk in 109.5\,ms, faster than
\pizerofive (120.5\,ms), $\Psi_0$ (149.7\,ms), and Fast-WAM (155.9\,ms), even
though Fast-WAM skips future generation at test time while \method retains
online foresight. GR00T N1.7 runs in 60.0\,ms using four denoising steps
without future generation. At our 10\,Hz control rate, a 30-step chunk spans
3\,s of execution for 110\,ms of model compute.

%% file: sec/5_conclusion.tex
\section{Conclusion}
\label{sec:conclusion}

We presented \method, a unified vision-language-action and latent world model.
By jointly denoising future spatial tokens and continuous actions in the VLM's
co-evolving perceptual space, \method unifies semantic reasoning, physical
foresight, and control without video generators or external encoders. Across
simulation and real-world humanoid tasks, \method substantially improves
whole-body loco-manipulation over VLA and WAM baselines at high efficiency, and
controlled interventions confirm that the policy actively relies on its
predicted dynamics. More broadly, our findings show that a VLM's perceptual
representations, when adapted jointly with foresight and control, provide an
effective substrate for embodied world modeling.

\noindent\textbf{Limitations.}
The foresight stream currently predicts the
future from the observation alone. Conditioning it on candidate actions would let \method compare
the outcomes of different actions and open the door to test-time planning in its own representation
space.

%% file: sec/XX_acknowledgement.tex
\section*{Acknowledgement}
\label{sec:acknowledgement}

We are grateful to our friends and colleagues, including Hamza Derbas, Kevin Huang, Jason Pries, Songlin Wei, Hongyi Jing, Hongyang Du, and others from Maven Robotics and USC Physical Superintelligence Lab, for their tremendous support, helpful discussions, and insightful suggestions.

%% file: sec/X_suppl.tex
\counterwithin{figure}{section}
\counterwithin{table}{section}
\counterwithin{equation}{section}
\raggedbottom
\setcounter{secnumdepth}{2}

\vspace{0.6em}
\etocdepthtag.toc{mtappendix}
\etocsettagdepth{mtmain}{none}
\etocsettagdepth{mtappendix}{subsection}
\etocsettocstyle{%
  \noindent{\large\sffamily\bfseries Contents}\par\vspace{4pt}%
  \noindent\rule{\linewidth}{0.5pt}\par\vspace{4pt}%
}{%
  \vspace{4pt}\noindent\rule{\linewidth}{0.5pt}\par\vspace{1.0em}%
}
\tableofcontents

\input{appendix/implementation_appendix}
\input{appendix/data_processing}
\input{appendix/deployment}
\input{appendix/baselines_appendix}
\input{appendix/simulation_appendix}
\input{appendix/foresight_analysis_appendix}
\input{appendix/design_axes_appendix}

%% file: appendix/implementation_appendix.tex
\section{Architecture and Implementation Details}
\label{app:impl}

This section details the factorized space-time visual encoder
(Appendix~\ref{app:temporal-encoding}), shared model configuration and
stage-wise training objectives (Appendix~\ref{app:stage-training}), and
target-domain adaptation schedules (Appendix~\ref{app:target-opt}).

\subsection{Factorized Space-Time Visual Encoding}
\label{app:temporal-encoding}

Inspired by MEM~\citep{mem2026}, the causal video encoder $V_\phi$ reuses the
weights of the VLM's vision encoder $E_\phi$ and introduces no extra
parameters. Most ViT blocks process each frame independently, while every
fourth block replaces frame-wise attention with factorized space-time attention
reusing the block's query, key, and value projections. Let $\vq_{p,j}$,
$\vk_{p,j}$, and $\vv_{p,j}$ denote the query, key, and value at spatial
position $p$ in frame $j$. The block first mixes values causally across time at
each spatial location and then attends spatially within each frame over the
temporally mixed values
\begin{align}
  \tilde{\vv}_{p,j} &= \sum_{j' \le j} \operatorname{softmax}_{j'}\!\Big(
    \tfrac{1}{\sqrt{d}}\, (\mR^{\mathrm{t}}_{j}\vq_{p,j})^{\top} \mR^{\mathrm{t}}_{j'}\vk_{p,j'} \Big)\, \vv_{p,j'}, \\
  \vy_{p,j} &= \sum_{p'} \operatorname{softmax}_{p'}\!\Big(
    \tfrac{1}{\sqrt{d}}\, (\mR^{\mathrm{s}}_{p}\vq_{p,j})^{\top} \mR^{\mathrm{s}}_{p'}\vk_{p',j} \Big)\, \tilde{\vv}_{p',j},
\end{align}
where $\mR^{\mathrm{t}}_{j}$ is a 1D rotary embedding of the frame index,
$\mR^{\mathrm{s}}_{p}$ is the pretrained 2D spatial rotary embedding, $d$ is
the head dimension, and $\vy_{p,j}$ passes through the block's residual
connection and MLP (attention heads omitted for brevity). Because temporal
mixing is causal, the final frame aggregates the full clip, and passing only
its $P$ tokens per camera to the VLM preserves the single-image token budget.
Multiple cameras are encoded independently and tagged with learned view-role
embeddings.

\subsection{Model Configuration and Stage-Wise Training}
\label{app:stage-training}

Table~\ref{tab:impl} summarizes the shared model architecture and prediction
hyperparameters, and Table~\ref{tab:training_stages} specifies the objectives,
gradient paths, and visual-noise schedules across the three training stages.
We pre-train and mid-train for 200k and 300k optimizer steps, respectively, at
global batch size 1{,}024 on 32 H200 GPUs. The peak learning rate is
$5\times10^{-5}$ after 2k warm-up steps, with a $0.1\times$ multiplier on the
vision tower. During mid-training, the EMA target encoder uses decay 0.99. For
future-feature prediction, the noise floor is sampled as
$u=1-\operatorname{sigmoid}(0.8z-0.8)$ with $z\sim\gN(0,1)$, setting visual
noise to $\sigma=\max(\tau,u)$ in mid-training and post-training.

\begin{table}[!t]
\centering
\caption{\textbf{Shared architecture and prediction settings.}}
\label{tab:impl}
\small
\setlength{\tabcolsep}{10pt}
\begin{tabular}{@{}ll@{}}
\toprule
Item & Setting \\
\midrule
VLM backbone & Qwen3.5-2B~\citep{qwen35} \\
Foresight / action expert width & 1024 / 1024 \\
Parameters (VLM / foresight / action) & 2.22B / 0.48B / 0.48B (3.18B total) \\
Temporal encoding & Space-time attention every fourth ViT block \\
Future prediction target & Spatial tokens of the final future frame \\
Target normalization & Parameter-free per-token layer normalization \\
Future-feature loss weight $\lambda_v$ & 0.1 \\
Velocity denominator floor $\sigma_{\min}$ & 0.05 \\
Action noise $\tau$ & $\mathrm{Beta}(1.5,1)$, rescaled to $[0.001,1]$ \\
Inference schedule & 10 joint Euler steps ($\sigma=\tau$) \\
\bottomrule
\end{tabular}
\end{table}

\begin{table}[!t]
\centering
\caption{\textbf{Training stages and gradient routing.} F, A, and V denote FAST
cross-entropy, action flow matching, and future-feature prediction,
respectively. Future targets are always stop-gradient EMA features. During
mid-training only F updates the VLM, whereas during post-training A updates the
VLM while V remains insulated from it. Experts and modality projections are
trainable whenever present.}
\label{tab:training_stages}
\small
\setlength{\tabcolsep}{6pt}
\renewcommand{\arraystretch}{1.15}
\begin{tabular}{@{}>{\raggedright\arraybackslash}m{0.13\linewidth}
>{\raggedright\arraybackslash}m{0.24\linewidth}
>{\raggedright\arraybackslash}m{0.12\linewidth}
>{\raggedright\arraybackslash}m{0.24\linewidth}
>{\raggedright\arraybackslash}m{0.13\linewidth}@{}}
\toprule
Stage & Data & Losses & Trainable modules & Visual noise $\sigma$ \\
\midrule
Pre-training & EgoDex, EgoVerse & F & VLM and state projection & Not used \\
\addlinespace[3pt]
Mid-training & AgiBot World, Humanoid Everyday, 20\% human replay & F + A + $0.1$V & VLM via F, foresight via A and V, action expert via A & $\sigma=\max(\tau,u)$ \\
\addlinespace[3pt]
Post-training & Target-domain demonstrations & A + $0.1$V & VLM via A, foresight via A and V, action expert via A & $\sigma=\max(\tau,u)$ \\
\bottomrule
\end{tabular}
\end{table}

\subsection{Target-Domain Adaptation and Ablation Controls}
\label{app:target-opt}

Table~\ref{tab:adaptation} lists the adaptation schedules across all four
evaluation settings. All main models optimize $\gL_{\text{flow}}+\lambda_v\gL_{\text{view}}$
using AdamW with $(\beta_1,\beta_2)=(0.9,0.95)$, weight decay 0.01, and EMA
decay 0.999 for the vision teacher.We decay the learning
rate by $10\times$ with a cosine schedule. Warm-up lasts 500 steps for G1 and SIMPLE, 2k for
LIBERO, and 1k for RoboTwin. Physical G1 adaptation uses the collected
demonstrations with input resolutions of
$224\times288$ (RealSense) and $224\times384$ (ZED), a 78-D action space, and a
43-D state padded to 48. Across all ablation controls
in Table~\ref{tab:ablation}, variants share each benchmark's mid-trained
initialization, demonstrations, and optimization schedule, including trainable
vision and language backbones.

\begin{table}[!t]
\centering
\caption{\textbf{Target-domain adaptation schedules.} Batch sizes are global.
LR denotes peak expert learning rate, with a $0.1\times$ multiplier on the
vision tower and the listed multiplier on the language model (LM). History and
stride refer to the causal video encoder in frames and dataset timesteps.}
\label{tab:adaptation}
\small
\setlength{\tabcolsep}{8pt}
\begin{tabular}{@{}lrrccccc@{}}
\toprule
Setting & Steps & Batch & LR & LM scale & Action horizon & History & Stride \\
\midrule
Unitree G1 & 20k & 128 & $10^{-4}$ & $0.5\times$ & 30 & 3 & 10 \\
LIBERO & 80k & 128 & $5\times10^{-5}$ & $1.0\times$ & 16 & 2 & 8 \\
RoboTwin 2.0 & 100k & 256 & $10^{-4}$ & $0.5\times$ & 50 & 2 & 25 \\
SIMPLE & 50k & 256 & $10^{-4}$ & $0.5\times$ & 30 & 3 & 10 \\
\bottomrule
\end{tabular}
\end{table}

%% file: appendix/data_processing.tex
\section{Training Data and Preprocessing}
\label{app:data}

\subsection{Dataset Composition and Standardization}
\label{app:data-standard}

We convert every training source into a unified episodic format
(LeRobotDataset v3.0\footnote{https://huggingface.co/docs/lerobot/en/lerobot-dataset-v3}) 
storing raw multimodal streams including camera images with intrinsics and per-frame extrinsics,
proprioception, tracked hand keypoints, and wrist or end-effector poses
expressed in the camera frame whenever extrinsics are available. Signals absent
from a dataset or lost during tracking receive an explicit ignore marker, and
per-dataset normalization statistics are computed with ignored entries
excluded. Mapping raw signals into the unified state and action space occurs on
the fly at data-loading time, turning ignored entries into per-channel validity
masks in the training losses. Table~\ref{tab:data} summarizes all pre-training,
mid-training, and G1 post-training sources.

\begin{table}[!t]
\centering
\caption{\textbf{Training datasets across stages.} Hours are computed from
frame counts at the native 30\,Hz recording rate (10\,Hz for our G1
demonstrations). Human datasets use the 48-D hand representation in the camera
frame, whereas Humanoid Everyday uses it in the robot base frame.}
\label{tab:data}
\small
\setlength{\tabcolsep}{5pt}
\adjustbox{max width=\linewidth}{%
\begin{tabular}{@{}llllrl@{}}
\toprule
Source & Type & Embodiment & Cameras & Hours & Stages \\
\midrule
EgoDex~\citep{hoque2025egodex}             & human & hands & head & 829.4 & 1, 2 \\
EgoVerse-Mecka~\citep{egoverse2026}         & human & hands & head & 826.2 & 1, 2 \\
EgoVerse-Scale~\citep{egoverse2026}         & human & hands & head & 216.6 & 1, 2 \\
EgoVerse-Aria~\citep{egoverse2026}          & human & hands & head & 104.9 & 1, 2 \\
\midrule
AgiBot World, gripper~\citep{agibot2025}    & robot & dual-arm, grippers & head + 2 wrist & 2{,}615.7 & 2 \\
AgiBot World, dexterous~\citep{agibot2025}  & robot & dual-arm, dex.\ hands & head & 105.9 & 2 \\
Humanoid Everyday G1~\citep{zhao2025humanoideveryday} & robot & humanoid & head & 16.5 & 2 \\
Humanoid Everyday H1~\citep{zhao2025humanoideveryday} & robot & humanoid & head & 15.3 & 2 \\
\midrule
G1 demonstrations (ours)                    & robot & humanoid + SONIC & head & 6.8 & 3 \\
\bottomrule
\end{tabular}}
\end{table}

\subsection{Temporal Smoothing of Human Hand Trajectories}
\label{app:smooth}

Per-frame vision-based hand estimates in EgoDex and EgoVerse exhibit
millimeter-scale frame-to-frame jitter. Because FAST tokenizes action chunks
via discrete cosine transform (DCT) coefficients, high-frequency jitter creates
many small non-zero coefficients and inflates sequence length. We smooth each
48-D human trajectory along time using a Gaussian kernel with standard
deviation $\sigma_s=2$ steps normalized by the smoothed validity mask
\begin{equation}
  \tilde{x}_t = \frac{\sum_{j} g_{\sigma_s}(t-j)\, m_j x_j}{\sum_{j} g_{\sigma_s}(t-j)\, m_j},
\end{equation}
where $m_j\in\{0,1\}$ indicates valid entries, while invalid entries remain
masked in the loss. We load $3\sigma_s$ padding frames on each side of the
training window to avoid boundary artifacts, while keeping the current state
$\vs_t$ unsmoothed. Robot trajectories come from commanded joint angles and are
not smoothed.

Table~\ref{tab:smooth} evaluates smoothing on 1{,}024 paired 30-step chunks
from 45 EgoDex episodes and 3{,}072 chunks from 24 EgoVerse episodes (1{,}024
per subset). With our fitted tokenizer, smoothing shortens FAST sequences by
13.1\% on EgoDex and 20.9\% on EgoVerse on average (and by 18.9\% and roughly
32\% at the 99th percentile). The released FAST tokenizer exhibits comparable
relative reductions (13.1\% and 22.9\%), confirming that compression gains come
from removing jitter rather than vocabulary refitting alone. Median keypoint
displacement is only 1.3\,mm on EgoDex and 3.0\,mm on EgoVerse, with larger
99th-percentile displacements (58.8\,mm and 73.8\,mm) occurring only at peak
velocities of rapid reaching motions.

\begin{table}[!t]
\centering
\caption{\textbf{Effect of temporal smoothing ($\sigma_s=0\rightarrow2$) on
human action chunks.} Token counts are means per chunk. Values marked $\approx$
are EgoVerse quantiles averaged across its three subsets.}
\label{tab:smooth}
\small
\setlength{\tabcolsep}{10pt}
\begin{tabular}{@{}llcc@{}}
\toprule
Metric group & Metric & EgoDex & EgoVerse \\
\midrule
\multirow{3}{*}{Fitted tokenizer} & Tokens per chunk & 120 $\rightarrow$ 105 & 160 $\rightarrow$ 126 \\
 & Reduction (mean) & 13.1\% & 20.9\% \\
 & Reduction (99th pct.) & 18.9\% & $\approx$32.0\% \\
\midrule
\multirow{3}{*}{Released tokenizer} & Tokens per chunk & 124 $\rightarrow$ 107 & 169 $\rightarrow$ 130 \\
 & Reduction (mean) & 13.1\% & 22.9\% \\
 & Reduction (99th pct.) & 18.4\% & $\approx$31.6\% \\
\midrule
\multirow{3}{*}{Keypoint displacement (mm)} & Median & 1.3 & $\approx$3.0 \\
 & Mean & 5.4 & 8.0 \\
 & 99th pct. & 58.8 & $\approx$73.8 \\
\bottomrule
\end{tabular}
\end{table}

%% file: appendix/deployment.tex
\section{Real-World Humanoid Setup and Tasks}
\label{app:deploy}

\subsection{Hardware and SONIC Teleoperation-Deployment Pipeline}
\label{app:deploy-pipeline}

We deploy real-world policies on a Unitree G1 humanoid equipped with two
7-DoF Dex3-1 dexterous hands and a single egocentric head camera. The
default head-mounted Intel RealSense D435i provides a narrow field of view
suited to tabletop tasks (\texttt{Dog} and \texttt{Puffs}), whereas tasks
involving locomotion (\texttt{Chair}, \texttt{Book}, \texttt{Trash}, and
\texttt{Fruit}) require wider peripheral visibility and use a head-mounted ZED
2i camera. Each task uses the same camera during data collection and
evaluation.

Rather than outputting raw whole-body joint targets, high-level policies act
through SONIC~\citep{sonic}, a learned whole-body motion-tracking controller.
Each 78-D action comprises a latent SONIC motion token paired with joint
targets for both dexterous hands, while the 43-D proprioceptive state comprises
body joint positions and commanded hand targets. Demonstrations are collected
through the identical control interface by a teleoperator wearing a VR headset
whose tracked upper-body poses, hand commands, and locomotion inputs drive
SONIC at 50\,Hz, recording synchronized camera frames, proprioception, motion
tokens, and hand targets at 10\,Hz. During deployment, \method runs on a single
NVIDIA RTX 4090 GPU in synchronous closed-loop execution, streaming complete
30-step action chunks at 10\,Hz while the SONIC decoder tracks each token in a
50\,Hz balance and whole-body control loop.

\subsection{Evaluation Tasks, Scoring, and Stage-Wise Outcomes}
\label{app:tasks}

We collect 557 teleoperated G1 demonstrations in total, comprising 355
demonstrations across six evaluated tasks (49--70 per task,
Table~\ref{tab:tasks}) and 202 demonstrations across four additional household
skills (such as picking up a bottle, stacking plates, and pouring water) that
supply related manipulation priors during multi-task training. We selected the
six evaluation tasks for skill diversity and difficulty, spanning short
tabletop pick-and-place, object transport, articulated-object manipulation, and
locomotion coupled to contact. Within each task, episodes are partitioned at the
episode level using a deterministic hash (seed 42, nominal 10\% validation
allocation with the remainder used for training), and all methods are evaluated
from their final training checkpoint over 10 trials per task from matched
initial configurations.

As summarized in Table~\ref{tab:tasks}, the six evaluation tasks span short
tabletop pick-and-place (\texttt{Dog}, \texttt{Puffs}, 16--28\,s mean duration)
and long-horizon multi-stage loco-manipulation (\texttt{Chair}, \texttt{Book},
\texttt{Trash}, \texttt{Fruit}, 35--84\,s mean duration). Each stage carries
equal partial progress credit (0.5 for two-stage tasks and 0.25 for four-stage
tasks), and full task success requires completing all stages.
Table~\ref{tab:g1_outcomes} reports the complete trial-count distributions over
completed subgoals across all six tasks and five evaluated policies.

\begin{table}[!t]
\centering
\caption{\textbf{Real-world Unitree G1 task specifications.} Duration denotes
mean demonstration length.}
\label{tab:tasks}
\small
\setlength{\tabcolsep}{3pt}
\renewcommand{\arraystretch}{1.15}
\begin{tabular}{@{}lccl>{\raggedright\arraybackslash}m{\dimexpr\linewidth-145pt\relax}@{}}
\toprule
Task & Demos & Duration & Camera & Stages (progress credit) \\
\midrule
\texttt{Dog}   & 60 & 15.9\,s & RealSense & Place the toy dog on the box (0.5), contain the dog within the box (0.5) \\
\addlinespace[3pt]
\texttt{Puffs} & 49 & 27.7\,s & RealSense & Pick up the puffs box (0.5), turn around and place it in the drawer (0.5) \\
\addlinespace[3pt]
\texttt{Chair} & 70 & 35.4\,s & ZED 2i    & Push the chair under the table (0.5), reach target pose and return to standing (0.5) \\
\addlinespace[3pt]
\texttt{Book}  & 61 & 54.7\,s & ZED 2i    & Pick up and hold the book (0.5), walk and place it on the shelf (0.5) \\
\addlinespace[3pt]
\texttt{Trash} & 58 & 56.8\,s & ZED 2i    & Pick up bottle (0.25), pick up paper (0.25), throw bottle in bin (0.25), throw paper in bin (0.25) \\
\addlinespace[3pt]
\texttt{Fruit} & 57 & 83.7\,s & ZED 2i    & Pick up fruit plate (0.25), open fridge door (0.25), place plate inside (0.25), close door (0.25) \\
\bottomrule
\end{tabular}
\end{table}

\begin{table}[!t]
\centering
\caption{\textbf{Subgoal completion distributions on the Unitree G1.} Entries
give trial counts (out of 10) completing $(0, 1, 2)$ subgoals for two-stage
tasks and $(0, 1, 2, 3, 4)$ subgoals for four-stage tasks.}
\label{tab:g1_outcomes}
\small
\setlength{\tabcolsep}{4pt}
\adjustbox{max width=\linewidth}{%
\begin{tabular}{@{}lcccccc@{}}
\toprule
Method & \texttt{Dog} $(0,1,2)$ & \texttt{Puffs} $(0,1,2)$ & \texttt{Chair} $(0,1,2)$ & \texttt{Book} $(0,1,2)$ & \texttt{Fruit} $(0,1,2,3,4)$ & \texttt{Trash} $(0,1,2,3,4)$ \\
\midrule
Fast-WAM~\citep{fastwam} & $(3, 1, 6)$ & $(0, 4, 6)$ & $(4, 0, 6)$ & $(2, 3, 5)$ & $(4, 1, 0, 0, 5)$ & $(2, 1, 2, 1, 4)$ \\
\pizerofive~\citep{pi05} & $(3, 0, 7)$ & $(2, 3, 5)$ & $(6, 0, 4)$ & $(9, 1, 0)$ & $(9, 1, 0, 0, 0)$ & $(9, 0, 1, 0, 0)$ \\
GR00T N1.7~\citep{gr00tn17} & $(1, 0, 9)$ & $(4, 3, 3)$ & $(5, 2, 3)$ & $(4, 5, 1)$ & $(4, 3, 0, 0, 3)$ & $(2, 3, 2, 0, 3)$ \\
$\Psi_0$~\citep{psi02026} & $(2, 1, 7)$ & $(3, 2, 5)$ & $(4, 0, 6)$ & $(4, 3, 3)$ & $(4, 5, 0, 0, 1)$ & $(2, 2, 4, 0, 2)$ \\
\midrule
\method (ours) & $(0, 1, 9)$ & $(2, 2, 6)$ & $(1, 0, 9)$ & $(1, 3, 6)$ & $(1, 2, 0, 0, 7)$ & $(0, 1, 1, 1, 7)$ \\
\bottomrule
\end{tabular}}
\end{table}

%% file: appendix/baselines_appendix.tex
\section{Baselines and Inference Profiling}
\label{app:baseline}

\subsection{Baseline Adaptation Recipes and Compute Budgets}
\label{app:baseline-recipes}

In addition to the primary humanoid baselines in Section~\ref{sec:exp}, the
tabletop evaluations compare against $\pi_0$~\citep{pi0},
VLA-JEPA~\citep{vlajepa2026}, Motus~\citep{motus}, starVLA~\citep{starvla2026},
ABot-M0~\citep{abotm02026}, Xiaomi Robotics-0~\citep{xiaomirobotics02026},
X-WAM~\citep{xwam2026}, and Spatial Forcing~\citep{spatialforcing2026}. 
On the physical
Unitree G1 and in SIMPLE, all evaluated policies share the same demonstration
datasets, state-action interfaces, and whole-body control stacks. Adaptation
budgets and trainable modules are summarized in Table~\ref{tab:g1_budget}.

\noindent\textbf{$\Psi_0$.} We fine-tune the humanoid foundation model of
\citet{psi02026} using its official implementation
for 40{,}000 steps on G1 ($5.12\mathrm{M}$ samples) and 50{,}000 steps on
SIMPLE ($12.8\mathrm{M}$ samples), evaluating the final checkpoints under the
shared multi-task protocols.

\noindent\textbf{\pizerofive.} We fine-tune the released \pizerofive base
checkpoint with all parameters trainable and official OpenPI preprocessing for
20k steps at global batch size 128, using a cosine learning rate from
$2.5\times10^{-5}$ to $2.5\times10^{-6}$ and 1{,}000 warm-up steps. The action
horizon is 30, matching \method, with action projections re-initialized for the
78-D G1 action space.

\noindent\textbf{GR00T N1.7 and unfrozen-VLM control.} Following its official
adaptation recipe, we fine-tune the released GR00T N1.7 checkpoint by updating
the embodiment projector, action diffusion transformer, and vision-language
layer normalization while keeping the vision tower and language model frozen.
On G1, we train for 20k steps at batch size 128 with a cosine learning rate
from $10^{-4}$ to 0, 500 warm-up steps, state dropout 0.2, percentile
normalization, and official image augmentation. Because \method updates its vision and language backbones
during adaptation, we additionally train an unfrozen GR00T N1.7 control on
SIMPLE with its vision tower, language model, projector, diffusion transformer,
and state and action encoders all trainable under the same $50\mathrm{k}\times256$
budget (Table~\ref{tab:gr00t_unfrozen}). Unfreezing the VLM raises GR00T N1.7's
average success from 72.3\% to 73.9\%, while \method remains higher at every
difficulty level with an 8.0-point overall advantage (81.9\%).

\noindent\textbf{Fast-WAM and training compute.} We evaluate the Fast-WAM
architecture without test-time video generation, coupling a Wan2.2-TI2V-5B
video diffusion transformer with a 1024-wide action diffusion transformer
trained jointly on 33-frame clips at $224\times320$ resolution with min-max
normalization (action horizon 32, cosine learning rate from $10^{-4}$, weight
decay 0.01, and global batch size 192). Because Fast-WAM backpropagates through
a 5B video diffusion backbone over 33-frame video clips at every step, its
per-sample training cost is substantially higher than that of a VLA policy, so
fewer sample presentations correspond to comparable wall-clock GPU compute.
Accordingly, we set Fast-WAM's step budgets to match \method's training
GPU-hours on both real-world G1 ($7.5\mathrm{k}\times192=1.44\mathrm{M}$
samples) and SIMPLE ($35\mathrm{k}\times192=6.72\mathrm{M}$ samples). Training
losses converge in both settings, and in our preliminary results extending
training further did not improve downstream performance.

\begin{table}[!t]
\centering
\caption{\textbf{Adaptation budgets on SIMPLE and real-world G1.} Budgets are
shown as optimizer steps $\times$ global batch size $=$ sample presentations.
Trainable modules
list the default recipe.}
\label{tab:g1_budget}
\small
\setlength{\tabcolsep}{5pt}
\renewcommand{\arraystretch}{1.18}
\newcommand{\modcell}[1]{{\renewcommand{\arraystretch}{1.05}\begin{tabular}[c]{@{}l@{}}#1\end{tabular}}}
\adjustbox{max width=\linewidth}{%
\begin{tabular}{@{}lccl@{}}
\toprule
\multirow{2}{*}[-2.5pt]{Method} & \multicolumn{2}{c}{Budget (steps $\times$ batch $=$ samples)} & \multirow{2}{*}[-2.5pt]{Default trainable modules} \\
\cmidrule(lr){2-3}
& SIMPLE & Real-world G1 & \\
\midrule
\pizerofive~\citep{pi05} & -- & $20\mathrm{k}\times128=2.56\mathrm{M}$ & VLM, action expert, and projections \\
\addlinespace[3pt]
GR00T N1.7~\citep{gr00tn17} & $50\mathrm{k}\times256=12.8\mathrm{M}$ & $20\mathrm{k}\times128=2.56\mathrm{M}$ & \modcell{Projector, action transformer,\\VLM layer norm} \\
\addlinespace[3pt]
Fast-WAM~\citep{fastwam} & $35\mathrm{k}\times192=6.72\mathrm{M}$ & $7.5\mathrm{k}\times192=1.44\mathrm{M}$ & \modcell{Video and action transformers,\\state encoder} \\
\addlinespace[3pt]
$\Psi_0$~\citep{psi02026} & $50\mathrm{k}\times256=12.8\mathrm{M}$ & $40\mathrm{k}\times128=5.12\mathrm{M}$ & VLM, action expert, and projections \\
\midrule
\method (ours) & $50\mathrm{k}\times256=12.8\mathrm{M}$ & $20\mathrm{k}\times128=2.56\mathrm{M}$ & \modcell{VLM, foresight and action experts,\\projections} \\
\bottomrule
\end{tabular}}
\end{table}

\begin{table}[!t]
    \centering
    \caption{\textbf{GR00T N1.7 trainable-backbone control on SIMPLE.}
    Success rates (\%) averaged over three evaluation passes.}
    \label{tab:gr00t_unfrozen}
    \small
    \setlength{\tabcolsep}{10pt}
    \begin{tabular}{@{}lcccc@{}}
    \toprule
    Model & L0 & L1 & L2 & Avg. \\
    \midrule
    GR00T N1.7 (official, frozen VLM) & 78.1 & 71.9 & 66.9 & 72.3 \\
    GR00T N1.7 (unfrozen VLM) & 78.3 & 79.2 & 64.2 & 73.9 \\
    \midrule
    \method (ours) & \textbf{86.7} & \textbf{85.8} & \textbf{73.1} & \textbf{81.9} \\
    \bottomrule
    \end{tabular}
    \end{table}

\subsection{Deployment Latency Profiling}
\label{app:latency}

Table~\ref{tab:latency} measures in-process action-chunk computation on one RTX
4090 with $480\times640$ inputs, 100 warm-up calls, 100 timed calls, and CUDA
synchronization. GR00T N1.7 uses four denoising steps, whereas \pizerofive,
Fast-WAM, $\Psi_0$, and \method use ten. These deployment-level measurements
isolate model compute and exclude image acquisition, network transport, and
low-level whole-body control. Disabling \method's foresight stream reduces
model computation by approximately 45\,ms per chunk, measuring the incremental
cost of online latent foresight directly.

For the joint-versus-sequential comparison in Table~\ref{tab:joint-vs-seq}, we
time the same SIMPLE checkpoint on one idle RTX 4090 with three-frame
$360\times640$ clips and 30-step action chunks (20 warm-ups and 100
CUDA-synchronized calls per mode in separate processes, caching the VLM
prefix). Joint denoising (10 shared steps) averages $114.8\pm0.1$\,ms (median
114.8\,ms), whereas sequential denoising (10 future steps followed by 10 action
steps) averages $208.1\pm0.3$\,ms (median 208.0\,ms).

%% file: appendix/simulation_appendix.tex
\section{Simulation Benchmark Protocols and Detailed Results}
\label{app:simulation}

\subsection{SIMPLE Whole-Body Loco-Manipulation}
\label{app:simple}

SIMPLE~\citep{wei2026simple} evaluates whole-body humanoid loco-manipulation
through a 50\,Hz decoupled whole-body controller (36-D actions, 32-D
proprioceptive state). All methods train on the same 17-task dataset and are
evaluated across three levels of domain randomization, where L0 varies
materials and distractors, L1 adds lighting variation, and L2 (i.e., Hard) varies target
initial positions (10 episodes per task and level per pass). The dataset
comprises 12 teleoperation (Teleop) tasks and 5 motion-planned (MP) tasks.
Across all trained policies, we observe a substantial performance gap between
the Teleop and MP tasks, which we attribute to their distinct demonstration
distributions. For a meaningful comparison, our main experiments focus on the
12-task Teleop suite, averaged over three evaluation passes of one checkpoint
per method (1{,}080 episodes total, measuring evaluation variability). For
computational tractability across multiple trained variants, the matched
mechanism ablations and interventions on SIMPLE in Tables~\ref{tab:ablation},
\ref{tab:prediction-designs}, \ref{tab:content-interventions},
and~\ref{tab:joint-vs-seq} evaluate one pass on the 12 Teleop tasks at L2 (10
episodes per task, 120 episodes per entry). For
completeness, Tables~\ref{tab:simple-full} and~\ref{tab:simple-full-tasks}
report aggregate and per-task success rates across both Teleop and MP suites
for all 17 tasks.

\begin{table}[!t]
\centering
\caption{\textbf{Full SIMPLE benchmark summary.} Success rates (\%) averaged
over the 12 Teleop tasks, 5 MP tasks, and all 17 tasks.}
\label{tab:simple-full}
\small
\setlength{\tabcolsep}{10pt}
\begin{tabular}{@{}lccc@{}}
\toprule
Method & Teleop (12) & MP (5) & All 17 \\
\midrule
GR00T N1.7~\citep{gr00tn17} & 72.3 & 30.0 & 59.9 \\
Fast-WAM~\citep{fastwam} & 66.0 & 25.3 & 54.1 \\
$\Psi_0$~\citep{psi02026} & 65.8 & 20.0 & 52.4 \\
\midrule
\method (ours) & \textbf{81.9} & \textbf{52.0} & \textbf{73.1} \\
\bottomrule
\end{tabular}
\end{table}

\begin{table}[!t]
\centering
\caption{\textbf{Per-task SIMPLE results across all 17 tasks.} Success rates
(\%) averaged over L0, L1, and L2 (10 episodes per level).}
\label{tab:simple-full-tasks}
\small
\setlength{\tabcolsep}{6pt}
\adjustbox{max width=\linewidth}{%
\begin{tabular}{@{}llcccc@{}}
\toprule
Suite & Task & \method (ours) & GR00T N1.7~\citep{gr00tn17} & Fast-WAM~\citep{fastwam} & $\Psi_0$~\citep{psi02026} \\
\midrule
Teleop & \texttt{BendHandoverTeleop} & 78.9 & 83.3 & \textbf{85.6} & 66.7 \\
Teleop & \texttt{BendPickAndPlaceTeleop} & \textbf{72.2} & 66.7 & 40.0 & 52.2 \\
Teleop & \texttt{BendPickTeleop} & \textbf{87.8} & 85.6 & 75.6 & 72.2 \\
Teleop & \texttt{CloseDoorTeleop} & \textbf{98.9} & 95.6 & 97.8 & \textbf{98.9} \\
Teleop & \texttt{HandoverTeleop} & \textbf{94.4} & 64.4 & \textbf{94.4} & 71.1 \\
Teleop & \texttt{LocomotionPickBetweenTablesTeleop} & \textbf{84.4} & 53.3 & 70.0 & 73.3 \\
Teleop & \texttt{OpenFaucetTeleop} & \textbf{71.1} & 58.9 & 10.0 & 53.3 \\
Teleop & \texttt{OpenOvenTeleop} & \textbf{84.4} & 83.3 & 63.3 & 67.8 \\
Teleop & \texttt{OpenTrashCanTeleop} & \textbf{84.4} & 76.7 & 72.2 & 48.9 \\
Teleop & \texttt{PickAndPlaceAndHugContainerTeleop} & \textbf{53.3} & 37.8 & 32.2 & 38.9 \\
Teleop & \texttt{PushOfficeChairTeleop} & \textbf{78.9} & 70.0 & 63.3 & 68.9 \\
Teleop & \texttt{XMovePickTeleop} & \textbf{93.3} & 92.2 & 87.8 & 77.8 \\
\midrule
MP & \texttt{BendPickAndPlaceOnSofaMP} & \textbf{30.0} & 10.0 & 3.3 & 3.3 \\
MP & \texttt{BendPickMP} & \textbf{90.0} & 76.7 & 43.3 & 13.3 \\
MP & \texttt{PickAndBendPlaceMP} & 26.7 & \textbf{56.7} & 0.0 & 20.0 \\
MP & \texttt{TabletopGraspMP} & \textbf{80.0} & 3.3 & 56.7 & 60.0 \\
MP & \texttt{TabletopHandoverMP} & \textbf{33.3} & 3.3 & 23.3 & 3.3 \\
\midrule
Teleop & Average (12 tasks) & \textbf{81.9} & 72.3 & 66.0 & 65.8 \\
MP & Average (5 tasks) & \textbf{52.0} & 30.0 & 25.3 & 20.0 \\
All & Average (17 tasks) & \textbf{73.1} & 59.9 & 54.1 & 52.4 \\
\bottomrule
\end{tabular}}
\end{table}

\subsection{LIBERO and RoboTwin 2.0 Tabletop Protocols}
\label{app:tabletop-protocols}

\noindent\textbf{LIBERO.} We jointly train a single \method policy across all
40 tasks from the four LIBERO suites (\texttt{Spatial}, \texttt{Object},
\texttt{Goal}, and \texttt{Long}) for 80k steps at global batch size 128
(Table~\ref{tab:adaptation}). Evaluation uses 50 episodes per task (500 per
suite, 2{,}000 episodes per variant).

\noindent\textbf{RoboTwin 2.0.} Following the official clean-only RGB protocol,
we jointly train a single \method policy across all 50 bimanual tasks using 50
clean demonstrations per task (2{,}500 total) without measured depth, executing
25 actions before replanning. The main benchmark comparison in
Table~\ref{tab:robotwin} evaluates 100 episodes per task under clean (C2C) and
randomized (C2R) conditions (5{,}000 episodes per condition). For
computational tractability across multiple trained variants, the matched
mechanism ablations and interventions in Tables~\ref{tab:ablation},
\ref{tab:prediction-designs}, and~\ref{tab:content-interventions} evaluate 20
episodes per task and condition (1{,}000 episodes per condition), where the
full model scores 82.40\% on C2C and 40.10\% on C2R.

%% file: appendix/foresight_analysis_appendix.tex
\section{Extended Foresight Quality and Intervention Analyses}
\label{sec:foresight-analysis}

\subsection{Held-Out Prediction Quality and Offline Action-Error Protocol}
\label{app:heldout-quality}

For the offline evaluation in Table~\ref{tab:prediction-quality-main}, we test
the deployed physical G1 checkpoint and its EMA target encoder on episode-level
held-out trajectories across all six G1 tasks. Out of 12{,}480 candidate
observation-to-future windows, 804 extend past the end of an episode without
valid future frames and are excluded, leaving 11{,}676 valid windows. Across
individual tasks, the predicted-over-present similarity gain
$\cos(\alpha, y) - \cos(\beta, y)$ is positive on every one of the six tasks,
with $\alpha - \beta$ capturing both object displacement and egocentric
head-camera motion.

To isolate how future-token accuracy affects action generation on the same
held-out windows, we compute normalized open-loop action-chunk MSE against
demonstration chunks while holding the initial Gaussian action-noise sample
fixed across conditions and varying only the future-feature stream supplied to
the action expert (Table~\ref{tab:offline-action-mse-summary}). Noising
ground-truth EMA future tokens $y$ to the shared denoising level $\sigma=\tau$
reduces action MSE on every task individually, whereas supplying clean
present-frame tokens $\beta$ ($\sigma=0$) raises action MSE by 28.3\%.

\begin{table}[!t]
\centering
\caption{\textbf{Offline action-chunk error under controlled future-feature
inputs.} Normalized MSE against held-out G1 demonstration chunks with shared
initial action noise.}
\label{tab:offline-action-mse-summary}
\small
\setlength{\tabcolsep}{10pt}
\begin{tabular}{@{}lcc@{}}
\toprule
Future-feature input to action expert & Normalized action MSE $\downarrow$ & Relative change \\
\midrule
Model's own prediction $\alpha$ ($\sigma=\tau$) & 0.0495 & Reference \\
Ground-truth future $y$, noised to $\sigma=\tau$ & \textbf{0.0436} & $-11.9\%$ \\
Clean present frame $\beta$ ($\sigma=0$) & 0.0635 & $+28.3\%$ \\
\bottomrule
\end{tabular}
\end{table}

\subsection{Additional Inference Interventions on Physical G1 and LIBERO}
\label{app:extended-interventions}

Complementing the SIMPLE Hard and RoboTwin 2.0 interventions in
Table~\ref{tab:content-interventions}, Table~\ref{tab:extended-interventions}
reports test-time interventions on physical G1 \texttt{Dog} and LIBERO with
model weights held fixed (on \texttt{LIBERO-Long}, sequential denoising in
Table~\ref{tab:joint-vs-seq} scores 94.8\% vs.\ 96.6\% for joint denoising).
Replacing the predicted stream with Gaussian noise under the shared
$\sigma=\tau$ schedule lowers LIBERO average success from 97.70\% to 75.15\%,
with a 45.8-point drop on long-horizon \texttt{LIBERO-Long} (96.6\% to 50.8\%).
Declaring Gaussian noise unreliable via $\sigma=1$ preserves 96.50\% average
success on LIBERO and retains 60\% on physical G1 \texttt{Dog}, whereas
asserting a static future (\textbf{Copy}, $\sigma=\tau$) collapses success to
0\% on G1 \texttt{Dog} and 7.05\% on LIBERO.

\begin{table}[!t]
\centering
\caption{\textbf{Additional test-time interventions on physical G1 and LIBERO.}
Scores (\%) with fixed model weights on physical G1 \texttt{Dog} (10 trials)
and LIBERO (2{,}000 episodes across four suites, including 500 on
\texttt{Long}).}
\label{tab:extended-interventions}
\small
\setlength{\tabcolsep}{10pt}
\begin{tabular}{@{}lcccc@{}}
\toprule
& \multicolumn{2}{c}{Real-world G1 \texttt{Dog}} & \multicolumn{2}{c}{LIBERO} \\
\cmidrule(lr){2-3}\cmidrule(lr){4-5}
Variant & SR & Progress & Avg. SR & \texttt{Long} \\
\midrule
Model prediction ($\sigma=\tau$) & \textbf{90} & \textbf{95.0} & \textbf{97.70} & \textbf{96.6} \\
Gaussian noise ($\sigma=1$) & 60 & 60.0 & 96.50 & 95.4 \\
Gaussian noise ($\sigma=\tau$) & -- & -- & 75.15 & 50.8 \\
Present frame (Copy, $\sigma=\tau$) & 0 & 0.0 & 7.05 & 0.0 \\
\bottomrule
\end{tabular}
\end{table}

%% file: appendix/design_axes_appendix.tex
\section{Discussion of Related Predictive Robot Policies}
\label{app:design_axes}

While recent robotic policies increasingly incorporate future visual signals
beyond pixel reconstruction, they differ fundamentally in how future states are
represented and coupled to action generation. Implicit world-modeling methods
such as FLARE~\citep{flare2025} (adopted in GR00T N1.5~\citep{gr00tn15}) align
intermediate activations of an action diffusion transformer toward compact
action-aware embeddings via cosine loss. This shapes internal representations
during training without constructing an explicit spatial future stream that
participates in action denoising. Conversely, video-based WAMs~\citep{dreamzero,
motus, fastwam} and codec-based unified policies such as F1~\citep{f1vla2025}
and InternVLA-A1~\citep{internvlaa1} generate discrete tokens or continuous
latents of a separate pre-trained image or video tokenizer. Even when explicit
pixel decoding is omitted along the action path at test time, future generation
operates in a separate generative codec space rather than the visual
representation space that grounds VLM perception.

Among policies that explicitly predict latent visual features, existing designs
rely on frozen target encoders. DreamVLA~\citep{dreamvla2025} supervises world
queries with dynamic regions, depth, and semantic features from frozen external
teachers and omits its prediction decoders at deployment, while
VLA-JEPA~\citep{vlajepa2026} predicts latent states against a frozen external
V-JEPA~2 video encoder.
Closer in architecture is UniJEPA~\citep{unijepa2025}, which also attaches a
feature-prediction expert and an action expert to a PaliGemma VLM in a
mixture-of-transformers, yet differs from \method across target space,
prediction objective, and action coupling. First, UniJEPA predicts future features in the space of a frozen target encoder
(an initial SigLIP copy or DINOv3), while fine-tuning the policy's own vision
encoder on robot data. Its action expert therefore reads current observations
and predicted futures from two different spaces that drift apart during
training. Second, its predictor uses deterministic MSE regression, collapsing
multimodal futures toward their conditional mean rather than modeling their
distribution. Third, its noise-free foresight stream is computed once in a
prefix pass and cached prior to action
denoising. \method differs on all three axes: an online EMA of the policy's own
causal video encoder keeps current observations and predicted futures in one
co-evolving feature space, foresight tokens are generated via conditional flow
matching, and foresight and continuous actions are denoised jointly across
shared Euler steps.